\documentclass[journal]{IEEEtran}
\usepackage{amsmath,graphicx}
\usepackage{booktabs}
\usepackage{cite}
\usepackage{multirow}
\usepackage{multicol}
\usepackage{tabularx}
\usepackage{float}
\usepackage{makecell}
\usepackage{subfigure}
\usepackage{enumitem}
\usepackage{amsfonts,amssymb}
\usepackage[colorlinks]{hyperref}
\usepackage{cleveref}

\crefname{figure}{Fig.}{Figs.}
\Crefname{figure}{Figure}{Figures}
\crefname{section}{Sect.}{Sects.}
\Crefname{section}{Section}{Section}
\crefname{equation}{Eq.}{Eqs.}
\Crefname{equation}{Equation}{Equation}
\crefname{table}{Table}{Tables.}
\Crefname{table}{Table}{Tables}

\ifCLASSINFOpdf
\else
\fi
\begin{document}
%
\title{PSGAN: A Generative Adversarial Network for Remote Sensing Image Pan-sharpening}
%
%
%

\author{Qingjie~Liu,~\IEEEmembership{Member,~IEEE,}
	    Huanyu~Zhou,
	    Qizhi~Xu,
        Xiangyu~Liu,
        and~Yunhong~Wang,~\IEEEmembership{Fellow,~IEEE}
}

\markboth{IEEE Transactions on Geoscience and Remote Sensing,~Vol.~14, No.~8, March~2020}%
{Shell \MakeLowercase{\textit{et al.}}: Bare Demo of IEEEtran.cls for IEEE Journals}
%



\maketitle

\begin{abstract}
This paper addresses the problem of remote sensing image pan-sharpening from the perspective of generative adversarial learning. We propose a novel deep neural network based method named PSGAN. To the best of our knowledge, this is one of the first attempts at producing high-quality pan-sharpened images with GANs. The PSGAN consists of two components: a generative network (i.e., generator) and a discriminative network (i.e., discriminator). The generator is designed to accept panchromatic (PAN) and multispectral (MS) images as inputs and maps them to the desired high-resolution (HR) MS images and the discriminator implements the adversarial training strategy for generating higher fidelity pan-sharpened images. In this paper, we evaluate several architectures and designs, namely two-stream input, stacking input, batch normalization layer, and attention mechanism to find the optimal solution for pan-sharpening. Extensive experiments on QuickBird, GaoFen-2, and WorldView-2 satellite images demonstrate that the proposed PSGANs not only are effective in generating high-quality HR MS images and superior to state-of-the-art methods and also generalize well to full-scale images.
\end{abstract}


%
\IEEEpeerreviewmaketitle

\section{Introduction}
\label{sec:intro}
\IEEEPARstart{R}{ecently}, a lot of high resolution (HR) optical Earth observation satellites, such as QuickBird, GeoEye, WorldView-2, and GaoFen-2 have been launched, providing researchers in remote sensing community a large amount of data available for various research fields, such as agriculture~\cite{mulla2013twenty}, land surveying~\cite{shalaby2007remote}, environmental monitoring~\cite{weng2009thermal}, and so on. To obtain better results, many of these applications require images at the highest resolution both in spatial and spectral domains. However, due to technical limitations~\cite{zhang2004understanding}, satellites usually carry two kinds of optical imaging sensors and acquire images at two different yet complementary modalities: one is a high resolution panchromatic (PAN) image and another one is a low resolution (LR) multi-spectral (MS) image. Pan-sharpening (i.e. panchromatic and multi-spectral image fusion), which aims at generating high spatial resolution MS images by combining spatial and spectral information of PAN and MS images, offers us a good solution to alleviate this problem. 

Pan-sharpening could be beneficial for many practical applications, such as change detection, land cover classification, so it has gained increasing attention within the remote sensing community. Many research efforts have been devoted to developing pan-sharpening algorithms during the last decades~\cite{thomas2008synthesis,xu2014high,vivone2015critical,ghassemian2016review,PNN,pannet,AIHS}. The most widely used approaches are so-called component substitution (CS) methods, popularized because of their easy implementation and low computation cost in practical applications~\cite{tu2001new,tu2004fast,AIHS}. The basic assumption of CS methods is that the geometric detail information of an MS image lies in its structural component that can be obtained by transforming it into a new space. Then the structural component is substituted or partially substituted by a histogram matched version of PAN to inject the spatial information. Finally, pan-sharpening is achieved after an inverse transformation. The PCA based~\cite{chavez1991comparison,shahdoosti2016combining}, the IHS based~\cite{tu2001new,xu2014high} and the Gram-Schmidt (GS) transform~\cite{laben2000process} based methods are those of the most widely known CS methods.

Another popular family is multi-resolution analysis (MRA) based methods. It has a well-known French name am$\acute{\mathrm{e}}$lioration de la r$\acute{\mathrm{e}}$solution spatiale par injection de structures (ARSIS)~\cite{ranchin2000fusion}, which means enhancement of the spatial resolution by structure injections. The MRA-based methods assume that the missing spatial information in MS can be inferred from the high frequency of the corresponding PAN image. To pan-sharpen an MS image, multi-resolution analysis algorithms, such as discrete wavelet transform (DWT)~\cite{pradhan2006estimation}, “$\grave{\mathrm{a}}$ trous” wavelet transform~\cite{nunez1999multiresolution} or curvelet transform~\cite{nencini2007remote} are applied on a PAN image to extract high-frequency information and then inject it into the corresponding MS image. 

In addition, pan-sharpening can be formulated as an inverse problem, in which PAN and MS images are considered as degraded versions of an HR MS image and it can be restored by resorting to some optimization procedures~\cite{chen2014image,he2014new,vivone2015pansharpening}. This is an ill-posed problem because much information has been lost during the degrading process. To obtain the optimal solution, regularizer~\cite{chen2014image} or prior knowledge~\cite{he2014new} are added into formulations. Or, pan-sharpening can be addressed from the perspective of machine learning. For instance, Li et al.~\cite{li2011new} and Zhu et al.~\cite{zhu2013sparse,zhu2016exploiting} modeled pan-sharpening from compressed sensing theory. Liu et al.~\cite{liu2014pan} addressed pan-sharpening from a manifold learning framework. 

\begin{figure*}[htb]
	\begin{minipage}{.24\linewidth}
		\centering
		\centerline{\includegraphics[width=4.4cm]{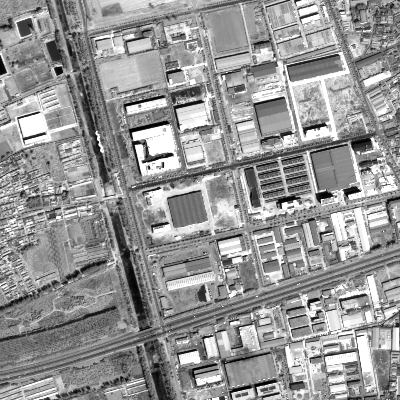}}
		\vspace{0.15cm}
		\centerline{\includegraphics[width=4.4cm]{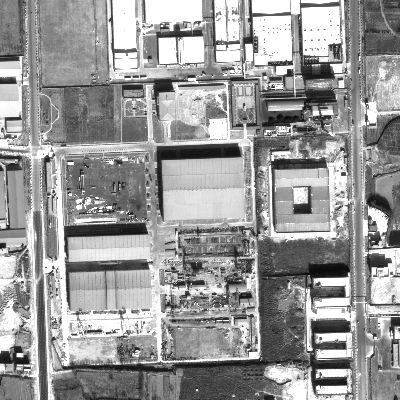}}
		\centerline{(a) PAN}
	\end{minipage}
	\hfill
	\begin{minipage}{.24\linewidth}
		\centering
		\centerline{\includegraphics[width=4.4cm]{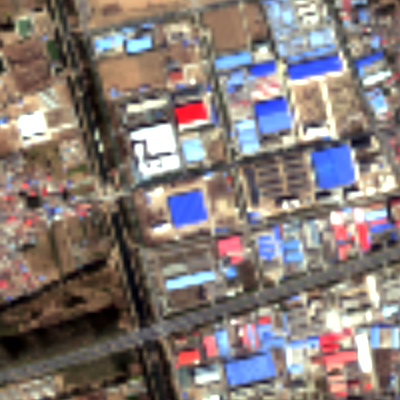}}
		\vspace{0.15cm}
		\centerline{\includegraphics[width=4.4cm]{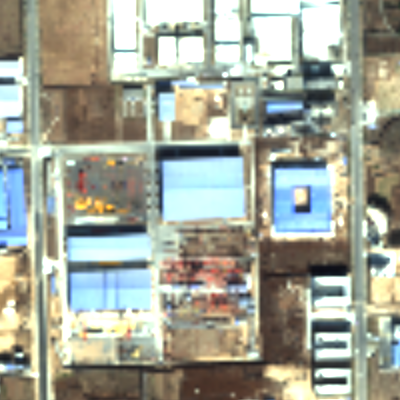}}
		\centerline{(b) LRMS}
	\end{minipage}
	\hfill
	\begin{minipage}{.24\linewidth}
		\centering
		\centerline{\includegraphics[width=4.4cm]{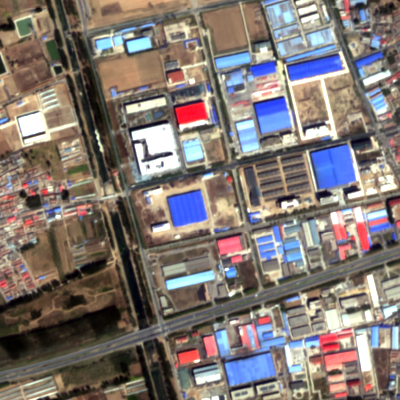}}
		\vspace{0.15cm}
		\centerline{\includegraphics[width=4.4cm]{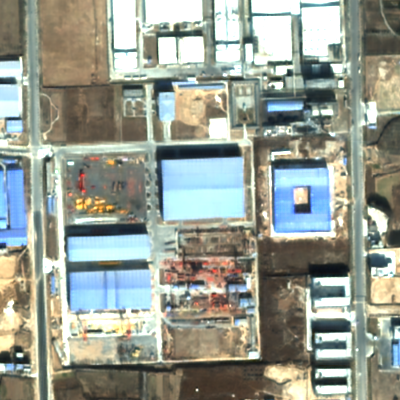}}
		\centerline{(c) Pan-sharpened}
	\end{minipage}
	\hfill
	\begin{minipage}{.24\linewidth}
		\centering
		\centerline{\includegraphics[width=4.4cm]{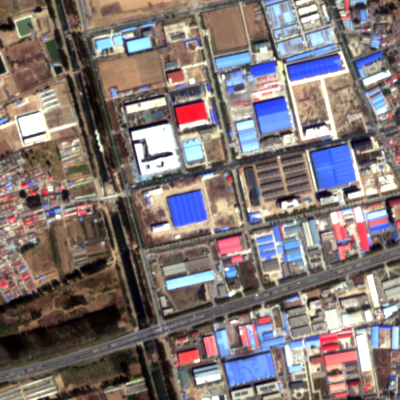}}
		\vspace{0.15cm}
		\centerline{\includegraphics[width=4.4cm]{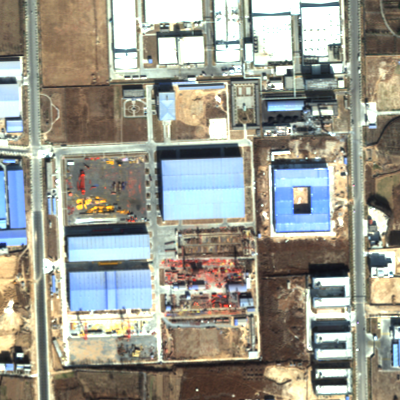}}
		\centerline{(d) Ground Truth}
	\end{minipage}
	\hfill	
	\caption{Example results of our PSGAN method. (c) are desired HR MS images generated from (a) PAN and (b) LR MS. (d) are ground truth HR MS images.}
	\label{fig:examples}
\end{figure*}

Recently, deep learning techniques have achieved great success in diverse computer vision tasks~\cite{dong2016image,zhang2016cnn,sreehari2017multi,ledig2017photo, yuan2020deep}, inspiring us to design deep learning models for the pan-sharpening problem. Observing that pan-sharpening and single image super-resolution share a similar spirit, and motivated by \cite{dong2016image}, Masi et al.~\cite{PNN} proposed a three-layered convolutional neural network (CNN) based pan-sharpening method and obtained improved results than traditional algorithms such as BDSD~\cite{garzelli2008optimal} and AWLP~\cite{wavelet}. Following this work, increasing attention has been paid to deep learning based pan-sharpening. For instance,  Zhong et al.~\cite{zhong2016remote} presented a CNN based hybrid pan-sharpening method. Different from \cite{PNN} that generate the pan-sharpened MS images directly, Zhong et al.'s work first enhances the spatial resolution of an input MS with the SRCNN method~\cite{dong2016image} and then applies GS transform on the enhanced MS and the PAN to accomplish the pan-sharpening. Rao et al.~\cite{rao2017residual} proposed CNN based pan-sharpening model built on top of SRCNN, in which SRCNN was employed to learn the difference between up-sampled MS image and ground truth. The final results were obtained by adding the predicted difference image to the up-sampled MS. Similarly, Wei et al.~\cite{wei2017boosting} proposed a much deeper network (11 layers) to learn the residual images.

Recent studies~\cite{szegedy2015going,DBLP:vgg,yuan2018multiscale} have suggested that deeper networks will achieve better performance on vision tasks. However, training becomes very difficult with depth increasing. Residual learning~\cite{he2016deep} ease this problem by introducing shortcut connections between different layers of a network, allowing training networks much deeper than previous ones. Pan-sharpening could also be improved by residual learning. Although, Rao et al.~\cite{rao2017residual} and Wei et al.~\cite{wei2017boosting} used the concept `residual network', the networks employed in their methods are built with plain units. The depth of their networks is still shallow. The first attempt at applying the residual network is PanNet~\cite{pannet}. They adopt a similar idea to \cite{rao2017residual} and \cite{wei2017boosting} but employ ResNet~\cite{he2016deep} to predict details of the image. In this way, both spatial and spectral information could be preserved well. 

Although great advances have been made in this field, there is still a great gap between the synthetic HR MS and the real one. It is still a challenging problem for researchers in the remote sensing community to obtain high spectral and spatial fidelity pan-sharpened images. To further boost the performance of pan-sharpening networks and obtain high-quality pan-sharpened images, in this paper we reformulate pan-sharpening as an image generation problem and explore the utilization of generative adversarial network (GAN)~\cite{goodfellow2014generative, 2020HPGAN} to solve it. The GAN framework is a powerful generative model and was first introduced by Goodfellow et al.~\cite{goodfellow2014generative}. In contrast to previous networks that have a unified architecture, GANs have two individual components: one generator that is trained to generate images indistinguishable from real ones, and one discriminator that tries to distinguish whether the generated images are real or fake. With this perspective, this paper proposes PSGAN, a generative adversarial network  that could produce high quality pan-sharpened images conditioned on the input of PAN and LR MS images.

This is an extension of our previous work~\cite{liu2018psgan}, which is the first work that addresses the pan-sharpening problem from the perspective of generative adversarial learning. Compared with \cite{liu2018psgan}, background knowledge about GAN is presented. And we give more details about the architecture of the proposed PSGAN and evaluate several possible architecture configurations of the PSGAN. We enlarge the dataset and conduct extensive experiments to demonstrate the effectiveness and superiority of it. The main contributions of this paper are as follows:
\begin{itemize}
	\item We address pan-sharpening problem from the perspective of image generation and develop novel generative adversarial networks for solving it.
	\item To accomplish pan-sharpening with the GAN framework, we design a basic two-stream CNN architecture as the generator to produce high-quality pan-sharpened images and employ a fully convolutional discriminator to learn adaptive loss function for improving the quality of the pan-sharpened images. 
	\item We evaluate various configurations of the proposed PSGAN and distill important factors that affect the performance of the pan-sharpening task.
	\item We demonstrate that the proposed PSGAN can produce astounding results on pan-sharpening problem. \textcolor{red}{Fig.~\ref{fig:examples}} shows one example result produced by our method. Codes are available~\footnote{https://github.com/zhysora/PSGan-Family}. 
\end{itemize}

The remainder of this paper is organized as follows: backgrounds and the theory of generative adversarial networks are briefly introduced in \Cref{sec:gan}. \Cref{sec:method} formulates pan-sharpening from the perspective of generative adversarial learning and gives details of proposed PSGAN architecture. Experiments are conducted in \Cref{sec:experiments}. And finally this paper is concluded in \Cref{sec:conclusion}.

\section{Generative Adversarial networks}
\label{sec:gan}
Given a set of unlabeled data, generative models aim at estimating their underlying distributions. This is a highly challenging task and inference on such distributions could be computationally expensive or even intractable. Recently proposed generative adversarial networks (GANs)~\cite{goodfellow2014generative} provide an efficient framework to learn generative models from unlabeled data. 

\begin{figure}
	\centering  
	\includegraphics[width=0.5\textwidth]{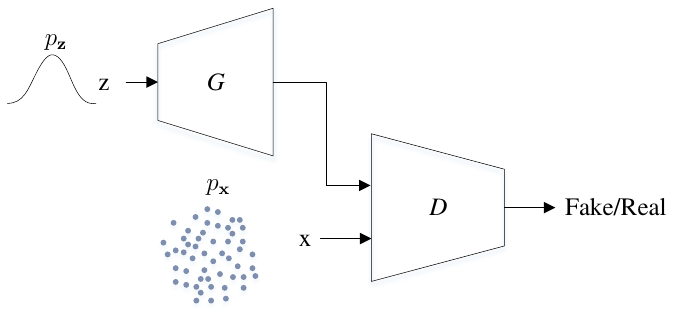} 
	\caption{An illustration of the generative adversarial framework. $G$ is a generator accepting a random signal $\mathbf{z}$ and trained to generate output that cannot be distinguished from a real data $\mathbf{x}$ by a discriminator $D$.}
	\label{fig:gan}  
\end{figure}

GANs learn generative models by setting up an adversarial game between a generator neural network $G$ and a discriminator neural network $D$. For any given data set $\{\mathbf{x}\}$, the generator $G$ learns the distribution of the data by mapping a random sample $\mathbf{z}$ from any distributions (e.g. Gaussian distribution or uniform distribution) to a sample $\mathbf{x}$ from the data space. The $G$ is trained to produce samples that can not be distinguished from the real samples. The discriminator $D$ output a scalar indicating the probability that the samples are produced by $G$ or it is from the real distribution. This process can be formulated as a two-player min-max game and written as follows:
\begin{equation}
\label{eq:gan}
\begin{aligned}
	\min\limits_G \max\limits_D V(D,G)&= \mathbb{E}_{\mathbf{x}\sim p_{\mathrm{data}}(\mathbf{x})}[\log D(\mathbf{x})]\\
	&+\mathbb{E}_{\mathbf{z}\sim p_{\mathrm{z}}(\mathbf{z})}[\log (1-D(G(\mathbf{z})))]
\end{aligned}
\end{equation}
where $p_{\mathrm{data}}(\mathbf{x})$ is the distribution of the real data, $\mathbf{x}$ is a sample from $p_{\mathrm{data}}(\mathbf{x})$. Correspondingly, $p_{\mathrm{z}}(\mathbf{z})$ is an arbitrary random distribution, and $\mathbf{z}$ is a sample drawn from it. The first term of Eq.~\ref{eq:gan} indicates the probability of the discriminator determines a sample is a `real' data, while the second term indicates the probability of the discriminator identify  a samples is `fake'. $D$ tries to assign correct labels to both real and generated data by maximizing the first term to 1 and the second term to 0. In contrast, $G$ takes a random noise $\mathbf{z}$ as input and tries to generate a sample that as indistinguishable from real one as possible by minimizing $\log (1-D(G(\mathbf{z})))$. An illustration of this procedure is given in \cref{fig:gan}.

Eq.~\ref{eq:gan} can be optimized in an iterative way by fixing one parameter and optimizing another one. When $G$ fixed, the optimization of $D$ can be considered as maximizing the log-likelihood of the conditional probability $p(Y=y|\mathbf{x})$, where $Y$ is the probability of sample $\mathbf{x}$ comes from the real data ($y=1$) or the fake data ($y=0$). When $D$ fixed, the objective of $G$ is minimizing the Jensen-Shannon divergence between the real data distribution $p_{\mathrm{data}}$ and the fake data distribution $p_G$ (here $p_G$ denotes distribution learned by the generator $G$). It can be proved that $G$ has an optimal solution $p_G = p_{\mathrm{data}}$~\cite{goodfellow2014generative}. And given enough capacity and training time, the generative neural network and the discriminator network will converge and achieve a point where the generator produce samples so real that the discriminator cannot distinguish them from the real data.

\section{PSGAN}
\label{sec:method}
\subsection{Formulation}
Pan-sharpening aims to estimate a pan-sharpened HR MS image $\hat{\mathbf{P}}$ from a LR MS image $\mathbf{X}$ and a HR PAN image $\mathbf{Y}$. The output images should be as close as possible to the ideal HR MS images $\mathbf{P}$. We describe $\mathbf{X}$ by a real-valued tensor of size $w\times h\times b$, $\mathbf{Y}$ by $rw\times rh\times 1$, and $\hat{\mathbf{P}}$ and $\mathbf{P}$ by $rw\times rh\times b$ respectively, where $r$ is the spatial resolution ratio between LR MS $\mathbf{X}$ and HR PAN $\mathbf{Y}$ (in this paper $r=4$) and $b$ is the number of bands. The ultimate goal of pan-sharpening takes a general form as follows:
{\setlength\abovedisplayskip{2pt}
	\setlength\belowdisplayskip{2pt}
	\begin{equation}\label{eq:generalform}
	\hat{\mathbf{P}} = f(\mathbf{X},\mathbf{Y};\Theta)
	\end{equation}}
where $f(\cdot)$ is a pan-sharpening model which takes $\mathbf{X}$ and $\mathbf{Y}$ as input and produces desired HR MS $\hat{\mathbf{P}}$, and $\Theta$ is collection of parameters for this model. \cref{eq:generalform} can be solved by minimizing the following loss function:
{\setlength\abovedisplayskip{2pt}
	\setlength\belowdisplayskip{2pt}
	\begin{equation}
	\label{eq:solve}
	\hat{\Theta}_f = \arg\min \sum\limits^N_{n=1} \ell  \left[f_{\Theta}(\mathbf{X}_n,\mathbf{Y}_n), \mathbf{P}_n  \right]
	\end{equation}
}
where $N$ is the number of training samples. As an example, \cref{eq:generalform} can be realized from the perspective of compressed sensing, and \cref{eq:solve} can be solved using dictionary learning algorithms~\cite{xie2013pan}.

From \cref{eq:generalform} we can see that $f(\cdot)$ can be considered as a mapping function from $(\mathbf{X}, \mathbf{Y})$ to $\mathbf{P}$. Thus, we can reformulate pan-sharpening as a conditional image generation problem that can be solved using conditional GAN~\cite{isola2017image}. Following~\cite{goodfellow2014generative} and~\cite{isola2017image}, we define a generative network $G$ that maps the joint distribution $p_{data}(\mathbf{X}, \mathbf{Y})$ to the target distribution $p_r(\mathbf{P})$. The generator $G$ tries to produce pan-sharpened image $\hat{\mathbf{P}}$ that cannot be distinguished from the reference image $\mathbf{P}$ by an adversarial trained discriminative network $D$. This can be expressed as a mini-max game problem:
{\setlength\abovedisplayskip{4pt}
	\setlength\belowdisplayskip{4pt}
	\begin{equation}
	\label{eq:ganloss}
	\begin{split}
	&\min\limits_{\Theta_G}\max\limits_{\Theta_D}\mathbb{E}_{\mathbf{X}\sim p_{data}(\mathbf{X}),\mathbf{P}\sim p_r(\mathbf{P})}[\log{D_{\Theta_D}(\mathbf{X},\mathbf{P})}]\\
	&+\mathbb{E}_{(\mathbf{X},\mathbf{Y})\sim p_{data}(\mathbf{X},\mathbf{Y})}[\log{(1-D_{\Theta_D}(\mathbf{X},G_{\Theta_G}(\mathbf{X},\mathbf{Y}))}].
	\end{split}
	\end{equation}}
With this adversarial learning, a GAN designed for pan-sharpening tasks could generate faithful HR MS images.

\begin{figure*}
	\centering  
	\includegraphics[width=\textwidth]{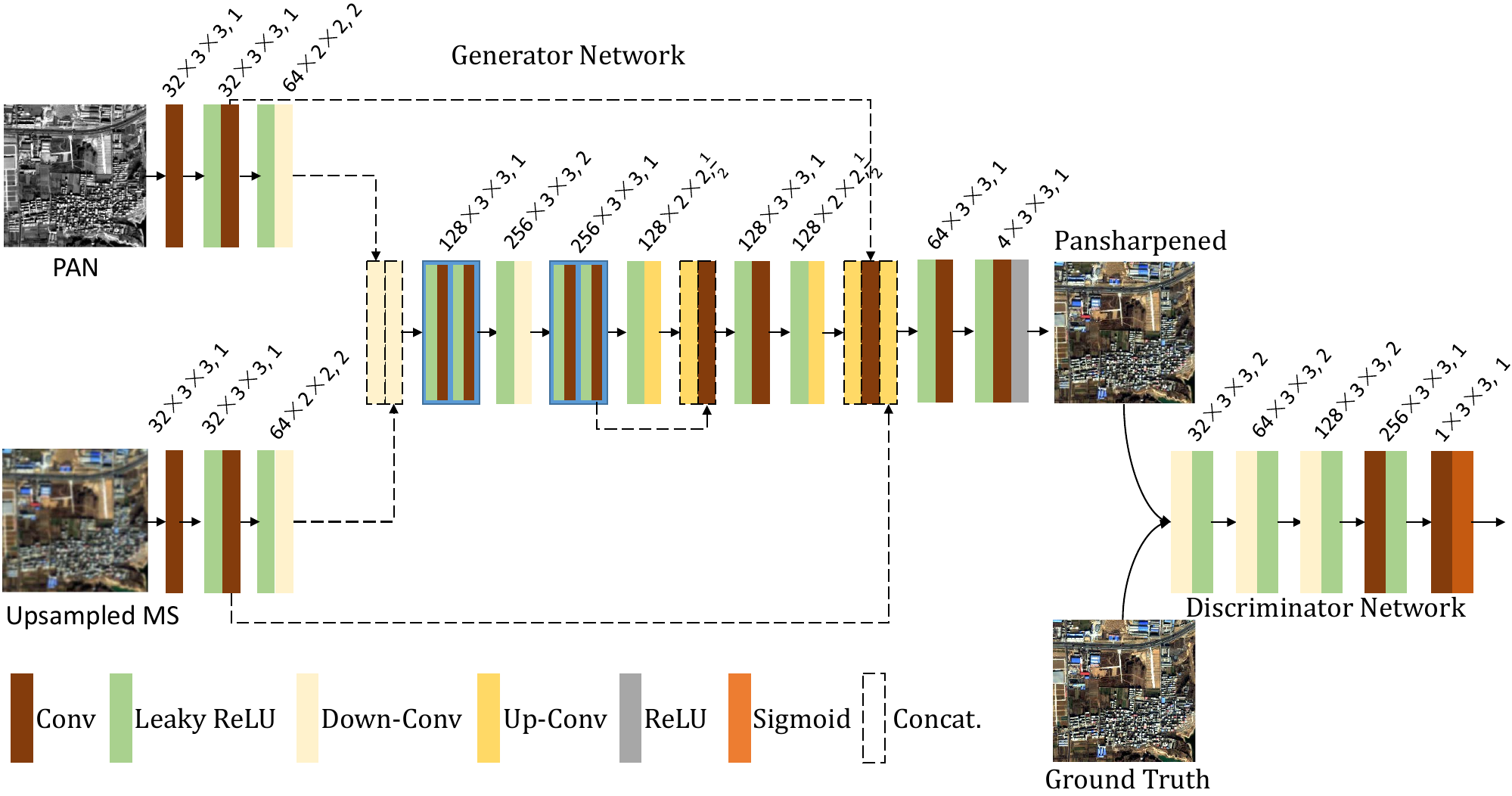} 
	\caption{Detailed architectures of the Generator network $G$ and the Discriminator network $D$.}
	\label{fig:architecture}  
\end{figure*}
\begin{figure} \centering	
	\subfigure[] { \label{generators:a}
		\includegraphics[width=0.49\textwidth]{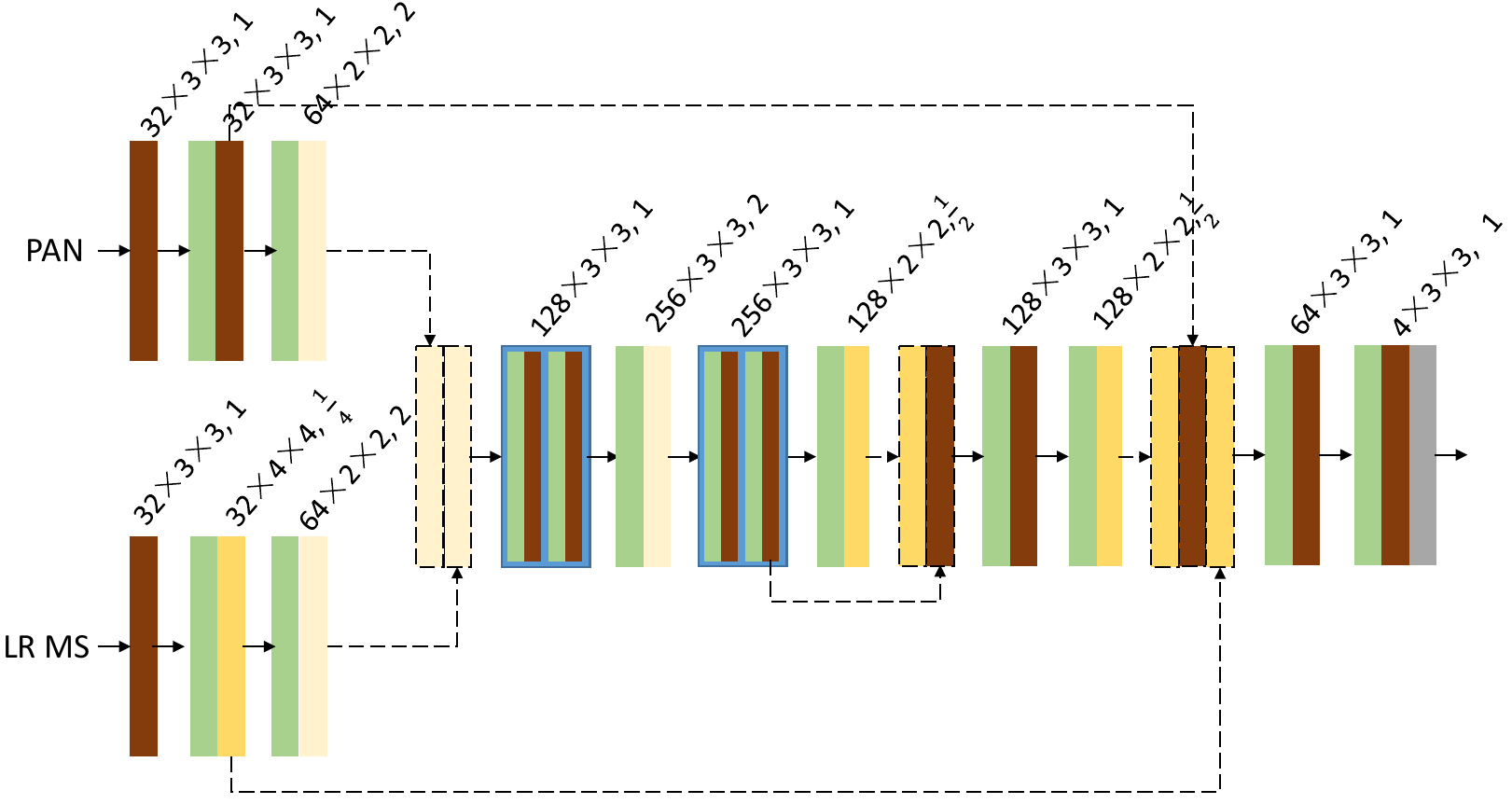}
	}\\
	\subfigure[] { \label{generators:b}
		\includegraphics[width=0.49\textwidth]{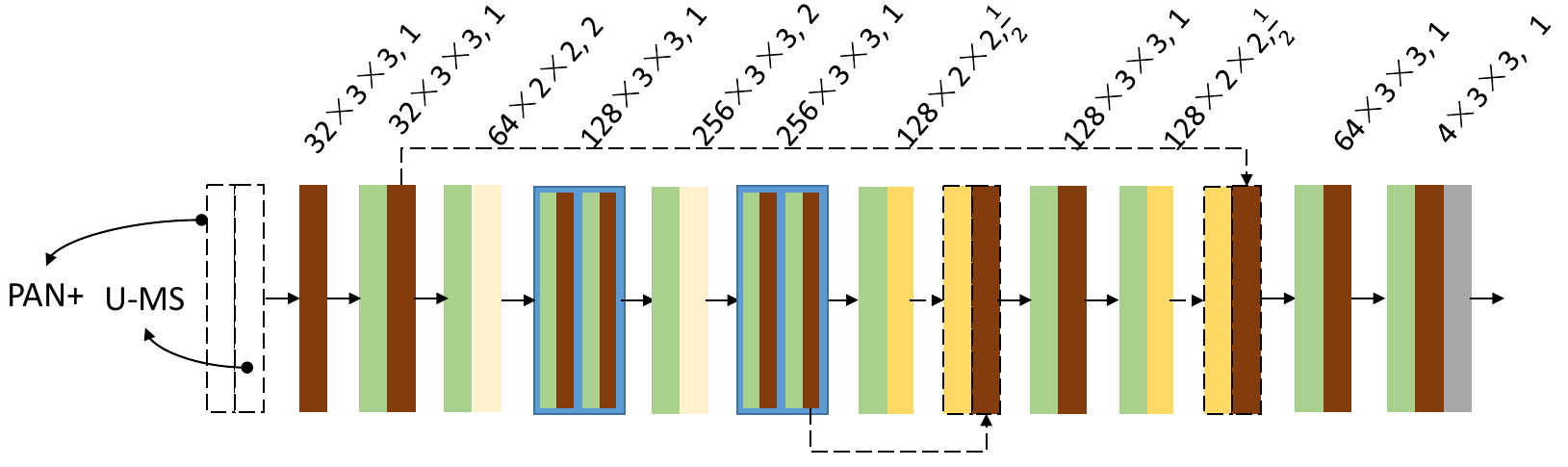}
	}
	\caption{Two variant generators for PSGANs. (a) is the generator performing up-scaling using networks; and (b) is the stacked generator that accepts one concatenated PAN and Upsampled MS (U-MS) image as input. Note: (a) and (b) share the same legend as \cref{fig:architecture}.}
	\label{fig:generators}
\end{figure}
\subsection{Architectures of the generator}
The ultimate goal of a generator is to produce a pan-sharpened MS image that cannot be distinguished from a real MS image. Since the inputs for a generator $G$ are a HR PAN image and a LR MS image, there are multiple ways to design the $G$. One possible way is using the network architecture similar to PNN~\cite{PNN} which stacks the PAN and the upsampled MS to form a five-band input~\footnote{In this paper, we only consider 4-band MS images.}. Another way is directly taking a two-stream design as in \cite{liu2020remote}. In this work, we devise and evaluate several generator architectures.
\subsubsection{Two-stream generator}
\label{subsubsec:twostreamG}
In contrast to other image generation tasks, e.g. single image super-resolution~\cite{Ledig_2017_CVPR}, image dehazing~\cite{Qu_2019_CVPR}, or face aging~\cite{yang2019learning}, that learn one-to-one mappings, pan-sharpening accepts two images acquired by different sensors with distinct characteristics over the same scene. The two modalities, i.e., the PAN image and the MS image contain different information. PAN image is the carrier of geometric detail (spatial) information, while MS image preserves spectral information. To make the best use of spatial and spectral information, we utilize two subnetworks to extract the hierarchical features of the input PAN and MS to capture complementary information of them. After that, the subsequent network proceeds as an auto-encoder: the encoder fuses information extracted from PAN and MS images, and the decoder reconstructs the HR MS images from the fused features in the final part. 

Considering that the spatial resolution of MS images is only $1/4$ of the desired pan-sharpened MS images, the pan-sharpening can be viewed as a special case of image super-resolution aided by PAN image. To enhance the spatial resolution of MS images using neural networks, there are usually two solutions. The \emph{first} one is upscaling MS images to the desired size using some interpolation methods such as bicubic, and then applying neural networks to learn nonlinear mapping. The \emph{second} one is applying the model directly without any preprocessing and performing up-scaling using networks. This will lead to deeper network structure and potentially better performance, however, with lower computational cost than the previous one~\cite{dong2016accelerating}. In this paper, we take into consideration both of the two solutions. 

\textbf{PSGAN} In our previous work~\cite{liu2018psgan}, we employ the first solution to build the PSGAN, which is up-sampling the MS image firstly and then feeding it and the corresponding PAN into two subnetworks for feature extraction. The architecture of the generator is shown in Fig.~\ref{fig:architecture}. The two subnetworks have a similar structure but different weights. Each of them consists of two successive convolutional layers followed by a leaky rectified linear unit (LeakyReLU)~\cite{leakyrelu} and a down-sampling layer. The convolutional layer with a stride of 2 instead of a simple pooling strategy e.g. max pooling is used to down-sample the feature maps. After passing through the two subnetworks, the feature maps are first concatenated and then fused by subsequent convolutional layers. Finally, a decoder-like network architecture comprised of 2 transposed convolutional and 3 flat convolutional layers is applied to reconstruct the desired HR MS images. Inspired by the U-Net~\cite{u-net}, we adapt the PSGAN network by adding skip connections. The skip connection will not only compensate details to higher layers but also ease the training. In the last layer, ReLU is used to guarantee the output is not negative. 

\textbf{FU-PSGAN} We take PSGAN as the base network and build variations on top of it. To differentiate different versions of PSGAN, we name the PSGAN with the second solution \mbox{FU-PSGAN} because the generator of it uses the \textbf{F}eature \mbox{\textbf{U}p-scaling} strategy. The generator of FU-PSGAN has almost identical architecture to PSGAN except that the MS subnetwork takes the original-sized MS as input and has one up convolution following the first convolution layer instead of a normal convolution layer, as shown in Fig.~\ref{generators:a}.  

\subsubsection{PAN \& MS Stacked generator}
Another possible way of designing generators is viewing the PAN and MS as a whole, i.e. stacking the two images along the channel dimension together to form a new image. To do this, the MS image should be up-sampled to match the size of the PAN image and then concatenated with the PAN to obtain an inflated image. One appealing advantage of this strategy is we can easily inherit some well-developed models from related research fields, such as single image super-resolution. For example, the pioneering PNN~\cite{PNN} borrows the main structure of the network from SRCNN~\cite{dong2016image}. 

\textbf{ST-PSGAN} Following PNN~\cite{PNN}, we design a deeper residual network to accomplish pan-sharpening. We call it \mbox{ST-PSGAN} since it has a \textbf{ST}acked generator. The generator of ST-PSGAN is shown in Fig.~\ref{generators:b}. It has almost the same structure to the generator of PSGAN in Fig.~\ref{fig:architecture} except for one major difference that one stream, along with the skip connection bound on it, is removed to be consistent with the stacked PAN and MS. Another imperceptible change is that the first convolution layer should adapt to the channel dimension of the new input. For fair comparisons, all three PSGANs share the same discriminator which will be described below.

\subsection{Fully convolutional discriminator}

In addition to the generator, a conditional discriminator network is trained simultaneously to discriminate the reference MS images from the generated pan-sharpened images. Similar to~\cite{isola2017image}, we use a fully convolutional discriminator, which consists of five layers with kernels of 3$\times$3. The stride of the first three layers is set to 2, and 1 for the last two layers. Except for the last layer, all the convolution layers are activated through LeakyReLU. Sigmoid is used to predict the probability of being \emph{real} HR MS or \emph{pan-sharpened} MS for each input. The architecture of the discriminator is shown in \cref{fig:architecture}.

We give the detailed parameters of the proposed PSGANs in \cref{fig:architecture,fig:generators}. Taking into account the trade-off between the model complexity and the performance, we do not build much deeper networks, although the depths of them can be deepened easily by inserting more convolution blocks. Also, larger kernels such as $5\times5$ or $7\times7$ are not considered in this work, because they bring much more parameters with the same network depth.

\subsection{Loss function}
We train the three models using the same loss function. In this subsection, we will take PSGAN as an example to describe the loss function. The generative network $G$ and the discriminator network $D$ are trained alternately. To optimize $G$ we adopt the pixel-wise loss and adversarial loss similar to some other state of the art GAN networks~\cite{isola2017image}. In contrast to many previous works~\cite{PNN,pannet} employing $\ell_2$ loss that calculates mean squared errors between the ground truth and the reconstructed images, in this work, we adopt $\ell_1$ loss that calculates the absolute difference between the pan-sharpened image and the ground truth. 
{\setlength\abovedisplayskip{2pt}
	\setlength\belowdisplayskip{2pt}
	\begin{equation}
	\label{g-loss}
	\begin{aligned}
	\mathcal{L}(G) = \sum_{n=1}^{N}[-&\alpha\log{D_{\Theta_D}(\mathbf{X},G_{\Theta_G}(\mathbf{X},\mathbf{Y}))}\\
	+&\beta\left \| \mathbf{P}-G_{\Theta_G}(\mathbf{X},\mathbf{Y}) \right \|_1]
	\end{aligned}
	\end{equation}}
Finally, the loss function for $D$ takes the form:
{\setlength\abovedisplayskip{2pt}
	\setlength\belowdisplayskip{2pt}
	\begin{equation}
	\label{d-loss}
	\begin{split}
	\mathcal{L}(D)=
	\sum_{n=1}^{N}[&1-\log{D_{\Theta_D}(\mathbf{X},G_{\theta_G}(\mathbf{X},\mathbf{Y}))}\\
	+&\log{D_{\Theta_D}(\mathbf{X},\mathbf{P})}]
	\end{split}.
	\end{equation}}
where $N$ is the number of training samples in a mini batch, $\alpha$, $\beta$ are hyper-parameters and are set to 1 and 100 in the experiments, respectively. 

\section{Experiments}
\label{sec:experiments}
In this section, we conduct extensive experiments to evaluate the effectiveness and superiority of the proposed PSGANs. 
\subsection{Dataset and implementation details}
We train and test our networks on three datasets comprised of images acquired by QuickBird (QB), GaoFen-2 (GF-2), and WorldView-2 (WV-2) satellite images. Since the desired HR MS images are not available, we follow Wald's protocol~\cite{Wald2000} to down-sample both the MS and PAN images with a factor of $r$ ($r=4$ in this paper). Then the original MS images are used as reference images to be compared with. We randomly crop patch pairs from the down-sampled MS and PAN to form training samples. It should be noted that we use larger patch size, $64\times 64 \times 4$ for MS patches and $256\times 256 \times 1$ for PAN patches, than our previous work~\cite{liu2018psgan}, in which the sizes for MS and PAN patches are $32\times 32 \times 4$ and $128\times 128 \times 1$, respectively. This will lead to smaller batch size during training, however, our experiments, which will be given in the next subsection, demonstrate that larger patch size produces better image quality. Brief information about the \textcolor{red}{three} datasets is illustrated in Table~\ref{tb:info_of_datasets}. All the results reported in the following subsections are based on the test sets which are independent of the training images.
\renewcommand\arraystretch{0.8}
\begin{table}[H]
	\centering
	\caption{Brief information about the three datasets used in experiments. Note: Spa. Res. means spatial resolution.}
	\label{tb:info_of_datasets}
	\begin{tabularx}{0.5\textwidth}{@{}llXX@{}}
		\toprule
		Dataset	& Images (Train/Test) & Training Samples  & Spa. Res. (PAN/MS) \\ 
		\midrule
		QB	    & 9 (8/1)             &25,038             &0.6/2.4 \\  
		\midrule
		GF-2	& 9 (8/1)             &13,460             &0.8/3.2\\  
		\midrule 
		WV-2	& 9 (8/1)             &11,552             & 0.5/2.0\\  
		\bottomrule
	\end{tabularx}
\end{table}
The PSGANs are implemented in PyTorch~\cite{paszke2019pytorch} and trained on a single NVIDIA Titan 2080Ti GPU. We use Adam optimizer~\cite{kingma2015adam} with an initial learning rate of 0.0002 and a momentum of 0.5 to minimize the loss function. The mini-batch size is set to 8.  It takes about 8 hours to train one model. 
The source codes and more experimental results are available at https://github.com/zhysora/PSGan-Family.
\subsection{Evaluation indexes}
We use five widely used metrics to evaluate the performance of the proposed and other methods on the four datasets, including SAM~\cite{Yuhas1992}, CC, sCC~\cite{zhou1998wavelet}, ERGAS~\cite{Wald2000}, and Q$_4$~\cite{wang2002Q}.

\begin{itemize}
	\item \textbf{SAM}
	The \emph{spectral angle mapper} (SAM)~\cite{Yuhas1992} measures spectral distortions of pan-sharpened images comparing with the reference images. It is defined as angles between the spectral vectors of pan-sharpened and reference images in the same pixel, which can be calculated as:
	\begin{equation}\label{equ:sam}
	\mathrm{SAM}(\mathrm{x}_1,\mathrm{x}_2)\triangleq\arccos\left(\frac{\mathrm{x}_1\cdot \mathrm{x}_2}{\parallel \mathrm{x}_1\parallel \cdot \parallel \mathrm{x}_2\parallel}\right)
	\end{equation}
	where $\mathrm{x}_{1}$ and $\mathrm{x}_{2}$ are two spectral vectors. SAM is averaged over all the images to generate a global measurement of spectral distortion. For the ideal pan-sharpened images, SAM should be 0.
	
	\item \textbf{CC}
	The \emph{Correlation Coefficient} (CC) is another widely used indicator measuring the spectral quality of pan-sharpened images. It calculates the CC between a pan-sharpened image $X$ and the corresponding reference image $Y$ as
	\begin{equation}\label{eq:cc}
	\mathrm{CC}\triangleq\frac{\sum\limits_{i=1}^{w}\sum\limits^{h}_{j=1}\left(X_{i,j}-\mu_{X}\right)\left(Y_{i,j}-\mu_{Y}\right)}{\sqrt{\sum\limits_{i=1}^{w}\sum\limits^{h}_{j=1}\left(X_{i,j}-\mu_{X}\right)^2\sum\limits^{w}_{i=1}\sum\limits^{h}_{j=1}\left(Y_{i,j}-\mu_{Y}\right)^2}}
	\end{equation}
	where $w$ and $h$ are the width and height of the images, $\mu_{*}$ indicates mean value of an image. CC ranges from -1 to +1, and the ideal value is +1.
	
	\item \textbf{sCC}
	To evaluate the similarity between the spatial details of pan-sharpened images and reference images, a high-pass filter is applied to obtain the high frequencies of them, then the correlation coefficient (CC) between the high frequencies is calculated. This quantity index is called \emph{spatial CC} (sCC)~\cite{zhou1998wavelet}. We use the high Laplacian pass filter given by,
	\begin{equation}\label{eq:hpass}
	F = \left[
	\begin{array}{ccc}
	-1 & -1 & -1 \\
	-1 & 8 & -1 \\
	-1 & -1 & -1
	\end{array}
	\right]
	\end{equation}
	to get the high frequency. A higher sCC indicates that most of the spatial information of the PAN image is injected during the fusion process. sCC is computed between each band of the pan-sharpened and reference image. The final sCC is averaged over all the bands of the MS images.
	
	\item \textbf{ERGAS}
	The \emph{erreur relative globale adimensionnelle de synth\`ese} (ERGAS), also known as the relative global dimensional synthesis error is a commonly used global quality index~\cite{Wald2000}. It is given by,
	\begin{equation}\label{eq:ergas}
	\mathrm{ERGAS} \triangleq 100\frac{h}{l}\sqrt{\frac{1}{N}\sum^N_{i=1}\left(\frac{\mathrm{RMSE}(B_i)}{M(B_i)}\right)^2}
	\end{equation}
	where $h$ and $l$ are the spatial resolution of PAN and MS images; RMSE($B_i$) is the root mean square error between the $i$th band of the fused and reference image; $M(B_i)$ is the mean value of the original MS band $B_i$.
	
	\item {$\mathbf{Q_4}$}
	The \emph{Quality-index} Q${_4}$~\cite{wang2002Q} is the 4-band extension of Q index~\cite{wald1997fusion}. Q${_4}$ is defined as:
	\begin{equation}
	\mathrm{Q_4}\triangleq\frac{4|\sigma_{\mathrm{z}_1\mathrm{z}_2}|\cdot|\mu_{\mathrm{z}_1}|\cdot|\mu_{\mathrm{z}_2}|}{(\sigma_{\mathrm{z}_1}^{2}+\sigma_{\mathrm{z}_2}^{2})({\mu_{\mathrm{z}_1}^2}+\mu_{\mathrm{z}_2}^2)}
	\end{equation}
	where $\mathrm{z}_1$ and $\mathrm{z}_2$ are two quaternions, formed with spectral vectors of MS images, i.e. $\mathrm{z} = a+\mathbf{\mathrm{i}}b+\mathbf{\mathrm{j}}c+\mathbf{\mathrm{k}}d$, $\mu_{\mathrm{z}_1}$ and $\mu_{\mathrm{z}_2}$ are the means of $\mathrm{z}_1$ and $\mathrm{z}_2$, $\sigma_{\mathrm{z}_1\mathrm{z}_2}$ denotes the covariance between $\mathrm{z}_1$ and $\mathrm{z}_2$, and $\sigma_{\mathrm{z}_1}^{2}$ and $\sigma_{\mathrm{z}_2}^{2}$ are the variances of $\mathrm{z}_1$ and $\mathrm{z}_2$.
	
\end{itemize}

Three non-reference metrics $D_{\lambda}$, $D_{S}$, and QNR are employed for full resolution assessment.

\begin{itemize}	
	\item $D_{\lambda}$~\cite{alparone2008multispectral} is a spectral quality indicator derived from the difference of inter-band $Q$ values calculated from the pan-sharpened MS
	bands and the low-resolution MS	bands. It is defined as:
	\begin{equation}\label{equ:dambda}
	D_{\lambda} \triangleq \sqrt{\frac{2}{K(K-1)}\sum_{i=1}^{K}\sum_{j=i}^{K}\left|Q(P_i,P_j)-Q(X_i,X_j)\right|}
	\end{equation}
	where $K$ is the number of bands for a MS image, $P_i$ and $X_i$ represent the $i$th band of the pan-sharpened and the LR MS images, respectively. 
	\item $D_{S}$~\cite{alparone2008multispectral} is a spatial quality metric complementary to $D_{\lambda}$. It is calculated as:
	\begin{equation}\label{equ:ds}
	D_{S} \triangleq \sqrt{\frac{1}{K}\sum_{i=1}^{K}\left|Q(P_i,Y)-Q(X_i,\tilde{Y})\right|}
	\end{equation}
	where $Y$ is a PAN image and $\tilde{Y}$ is its degraded low-resolution version. Both $D_{\lambda}$ and $D_S$ take values in [0,1], and the lower the better. 
	
	\item \textbf{QNR}~\cite{alparone2008multispectral} is the abbreviation of \emph{Quality with No \mbox{Reference}}. It is a combination of $D_{\lambda}$ and $D_S$ and measures global quality of fused images without any reference image. It is given by:
	\begin{equation}\label{equ:qnr}
	\mathrm{QNR} \triangleq (1-D_{\lambda})(1-D_S)
	\end{equation}
	The ideal value of QNR is 1.	
\end{itemize}
\subsection{Impact of patch size}
\begin{table*}[!htb]
	\centering
	\caption{Performance variations with respect to patch sizes. The batch size decreases from 64 to 8 with patch size increasing from 16 to 64. }
	\label{lable:patchsize}
	\begin{tabularx}{1\textwidth}{@{}lrXXXXXX@{}}
		\toprule
		&       & Patch Size & SAM $\downarrow$ & CC $\uparrow$ & sCC $\uparrow$  & ERGAS $\downarrow$ & Q4 $\uparrow$ \\ 
		\midrule
		\multirow{8}{*}{QB} 
		& \multirow{3}{*}{PSGAN} & 16 & 1.2514 & 0.9862 & 0.9867 & 1.3703 & 0.9852 \\
    	&	& 32  & 1.2270 & 0.9867 & 0.9869 &1.3594 & 0.9852 \\ 
		&                          & 64  & 1.1740 & 0.9877 & 0.9880 &1.2602 & 0.9869 \\ 
		\cmidrule(l){2-8} 
		&\multirow{3}{*}{FU-PSGAN} &16 & 1.3338 & 0.9855 & 0.9853 & 1.4167 & 0.9842 \\
		&	& 32  & 1.2883 & 0.9857 & 0.9861 & 1.3943 & 0.9845 \\
		&                          & 64  & 1.2411 & 0.9869 & 0.9865 & 1.2907 & 0.9864 \\
		\cmidrule(l){2-8} 
		& \multirow{3}{*}{ST-PSGAN}& 16 & 1.3114 & 0.9874 & 0.9864 & 1.3225 & 0.9865 \\
		&	& 32  & 1.3125 & 0.9867 & 0.9869 & 1.3662 & 0.9854 \\
		&                          & 64  & 1.2889 & 0.9869 & 0.9868 & 1.3267 & 0.9857 \\ 
		\midrule
		\multirow{8}{*}{GF-2} 
		& \multirow{3}{*}{PSGAN} & 16 & 0.7632 & 0.9905 & 0.9927 & 0.7388 & 0.9981\\
		&	& 32  & 0.7484 & 0.9908 & 0.9929 & 0.7303 & 0.9981 \\ 
		&                          & 64  & 0.7575 & 0.9909 & 0.9929 & 0.7233 & 0.9980 \\ 
		\cmidrule(l){2-8} 
		& \multirow{3}{*}{FU-PSGAN} & 16 & 0.7527 & 0.9908 & 0.9930 & 0.7287 & 	0.9981\\
		&	& 32  & 0.7456 & 0.9909 & 0.9931 & 0.7214 & 0.9982 \\
		&                          & 64  & 0.7181 & 0.9915 & 0.9935 & 0.7013 & 0.9982 \\
		\cmidrule(l){2-8} 
		& \multirow{3}{*}{ST-PSGAN} & 16 & 0.8325 & 0.9894 & 0.9914 & 0.7832 & 	0.9978\\
		&	& 32  & 0.7856 & 0.9903 & 0.9924 & 0.7477 & 0.9980\\
		&                          & 64  & 0.7300 & 0.9913 & 0.9933 & 0.7084 & 0.9981\\
		\bottomrule
	\end{tabularx}
\end{table*}
\begin{table}[!htb]
	\centering
	\caption{Impacts of batch normalization and self-attention on the four datasets. `+BN' means PSGAN is modified by adding batch normalization after each convolution block. `+SA' means PSGAN is equipped with self-attention.}
	\label{lable:differentconfigs}
	\begin{tabularx}{0.5\textwidth}{@{}llXXXXX@{}}
		\toprule
		&        & SAM$\downarrow$ & CC$\uparrow$ & sCC$\uparrow$ & ERGAS$\downarrow$ & Q$_4$$\uparrow$ \\ 
		\midrule
		\multirow{5}{*}{QB} 
		&PSGAN     & 1.1740 & 0.9877 & 0.9880 & 1.2602 & 0.9869 \\ 
		&PSGAN-f16 & 1.5760 & 0.9832 & 0.9818 & 1.4414 & 0.9832\\ 
		&PSGAN-k$5\times 5$ &1.2843 & 0.9845 & 0.9863 & 1.4316 & 0.9826 \\ 
		&PSGAN+BN  & 9.2521 & 0.9274 & 0.9075 & 14.029 & 0.8167 \\  
		&PSGAN+SA  & 2.1396 & 0.9696 & 0.9738 & 2.3140 & 0.9629 \\ 
		\midrule
		\multirow{5}{*}{GF-2} 
		&PSGAN     & 0.7575 & 0.9909 & 0.9929 &	0.7233 & 0.9980 \\ 
		&PSGAN-f16 & 0.8411 & 0.9889 & 0.9908 & 0.7994 & 0.9976 \\ 
		&PSGAN-k$5\times 5$  & 0.7293 & 0.9914 & 0.9932 & 0.7025 & 0.9981 \\ 
		&PSGAN+BN  & 1.5903 & 0.9669 & 0.9698 & 7.6059 & 0.9260 \\  
		&PSGAN+SA  & 1.2182 & 0.9739 & 0.9793 & 1.2422 & 0.9944 \\ 
		\midrule
		\multirow{5}{*}{WV-2} 
		&PSGAN     & 0.9127 & 0.9973 & 0.9975 &	1.6452 & 0.9971 \\
		&PSGAN-f16 &0.9955 & 0.9968 & 0.9970 & 1.7718 & 0.9966 \\ 
		&PSGAN-k$5\times 5$ & 1.0061 & 0.9972 & 0.9971 & 1.7002 & 0.9968 \\ 
		&PSGAN+BN  & 8.1520 & 0.8697 & 0.8448 & 9.6634 & 0.8016 \\  
		&PSGAN+SA  & 1.4185 & 0.9950 & 0.9943 & 2.2273 & 0.9944 \\ 
		\bottomrule
	\end{tabularx}
\end{table}
In our previous work~\cite{liu2018psgan}, we use small patch size to generate training samples, which allows us to set a larger batch size and thus enables more stable training and faster convergence~\cite{goyal2017accurate}. However, for image reconstruction tasks, larger patch size is beneficial for generating high-quality images. In this paper, we test a much larger patch size than \cite{liu2018psgan}. Although the batch size will decrease accordingly, our experiments demonstrate that larger patches lead to higher image quality. We conduct experiments on the QB and GF-2 images to evaluate how much the impact will be by setting different patch sizes. The results are given in Table~\ref{lable:patchsize}, from which we can see larger patch size does have a positive impact on the image quality. For all the three models, the image quality has a significant improvement when using the 64 patch size, even though the batch is decreased from 32 to 8. Especially, the spectral indicator SAM and the global quality measurement ERGAS have obvious superiority with one exception that the SAM for PSGAN has slightly lower value on GF-2 images. The effect of patch size on CC, sCC, and Q4 is weak, only with small improvements and sometimes even slightly worse. Even though, we are encouraged to use larger patch size since the spectral quality is very important in the pan-sharpening task. Thus, in the following experiments, the patch size is set as 64.

\subsection{Impact of number of feature maps and kernel size}
Kernel size and the number of feature maps are important factors when designing neural networks. We test two designs of our PSGAN to evaluate impacts of feature maps and kernel sizes. Firstly, we reduce the number of the feature maps by a half, thus leading to fewer parameters. Secondly, we replace the typical $3\times 3$ convolutional filters with $5\times 5$ kernels. Enlarging kernel size would increase the number of parameters of the model, dramatically, as shown in Table~\ref{lable:computationalcosts}. We name these two designs PSGAN-f16 and PSGAN-k$5\times 5$, respectively. Test results are given in Table~\ref{lable:differentconfigs}. Although PSGAN-f16 has fewer parameters (about 1/4 of PSGAN) and runs faster than other PSGANs, its performance is not satisfactory. It obtains better results than PSGAN+BN and PSGAN+SA, but weaker than PSGAN and PSGAN-k$5\times 5$. PSGAN-k$5\times 5$ is with more than $2\times$ parameters than PSGAN and about $10\times$ parameters than PSGAN-f16. Such huge parameters make networks hard to train. From Table~\ref{lable:differentconfigs} we can see, PSGAN-k$5\times 5$ works well even better than PSGAN on the GF-2 images, however, obtains worse results on the QB and WV-2 datasets. 
\subsection{Batch normalization is harmful}
Batch normalization (BN)~\cite{ioffe2015batch} has been widely used in neural networks to stabilize and accelerate training. It also has been applied to the pan-sharpening task for improving performance~\cite{scarpa2018target,dian2018deep}. However, recent studies have suggested that batch normalization may be unnecessary in low-level visions~\cite{lim2017enhanced}. It brings two burdens: firstly, BN operation requires an amount of storage and computational resources, which could be used to add more convolutional layers. Secondly, BN layers get rid of scale information, which is helpful for recognition tasks, however, is harmful to scale-sensitive tasks, such as image super-resolution and pan-sharpening. We add a BN layer into each block for comparison. The results are given in Table~\ref{lable:differentconfigs}. We can observe that adding BN layers severely decrease the performance, especially on the QB and WV-2 images. Thus, in this work, we remove all the BN layers from our models.

\begin{table*}[!htb]
	\centering
	\caption{Performance comparisons on the test set of QB.The top three performances are highlighted with \textbf{\textcolor{red}{Red}}, \textbf{\textcolor{green}{Green}}, and \textbf{\textcolor{blue}{Blue}}.}
	\label{table:compareonQB}
	\begin{tabularx}{1\textwidth}{@{}lrXXXXX|XXX@{}}
		\toprule
		&         & SAM$\downarrow$ & CC$\uparrow$ & sCC$\uparrow$ & ERGAS$\downarrow$ & Q4$\uparrow$ & $D_{\lambda}\downarrow$ & $D_S\downarrow$ & QNR$\uparrow$ \\ 
		\midrule
		&SFIM~\cite{SFIM} & 1.3465 & 0.9620 & 0.9752 & 2.6051 & 0.9643  &\textbf{\textcolor{blue}{0.0062}} & 0.0170 & 0.9769\\
		&LMVM~\cite{LMVM} & 1.7131 & 0.9694 & 0.9703 & 2.3509 & 0.9647  &\textbf{\textcolor{green}{0.0020}} & 0.0164 & \textbf{\textcolor{blue}{0.9816}}\\
		&LMM~\cite{LMVM} & 1.6845 & 0.9634 & 0.9695 & 2.4306 & 0.9640  &0.0064 & 0.0173 & 0.9763\\
		&HPF~\cite{HPF} & 1.3522 & 0.9699 & 0.9811 & 2.2534 & 0.9698  &0.0069 & 0.0178 & 0.9755\\
		&HPFC~\cite{HPF} & 1.6558 & 0.9609 & 0.9776 & 4.2814 & 0.9453  &0.0461 & 0.0468 & 0.9093\\
		&Brovey~\cite{brovey} & 1.4782 & 0.9729 & 0.9720 & 2.0542 & 0.9735  &0.0281 & 0.0503 & 0.9231\\
		&HCS~\cite{HCS} & 1.4782 & 0.9729 & 0.9685 & 2.5003 & 0.9632  &0.0137 & 0.0285 & 0.9582\\
		&IHS~\cite{IHS} & 1.6100 & 0.9683 & 0.9822 & 2.2611 & 0.9697  &0.0078 & 0.0550 & 0.9376\\
		&GS~\cite{GS} & 1.3063 & 0.9726 & 0.9821 & 2.1309 & 0.9704  &0.0232 & 0.0497 & 0.9283\\
		&BDSD~\cite{BDSD} & 1.4725 & 0.9725 & 0.9864 & 2.2722 & 0.9707  &0.0147 & 0.0227 & 0.9629\\
		\hline
		\hline
		&PNN~\cite{PNN} & 2.0777 & 0.9731 & 0.9718 & 1.8752 & 0.9723  & 0.0273 & 0.0278 & 0.9457  \\
		&PanNet~\cite{pannet} & \textbf{\textcolor{red}{1.1068}} & \textbf{\textcolor{blue}{0.9848}} & \textbf{\textcolor{green}{0.9877}} & 1.3800 & 0.9834  & \textbf{\textcolor{red}{0.0019}} & \textbf{\textcolor{red}{0.0111}} & \textbf{\textcolor{red}{0.9871}}  \\
		&RED-cGAN~\cite{shao2019residual} & 1.2541 & 0.9868 & 0.9867 & \textbf{\textcolor{blue}{1.2932}} & \textbf{\textcolor{blue}{0.9862}} & 0.0069 & 0.0183 & 0.9749 \\
		&PSGAN  & \textbf{\textcolor{green}{1.1740}} & \textbf{\textcolor{red}{0.9877}} & \textbf{\textcolor{red}{0.9880}} &\textbf{\textcolor{red}{1.2602}} & \textbf{\textcolor{red}{0.9869}} & 0.0067 & \textbf{\textcolor{green}{0.0116}} & \textbf{\textcolor{green}{0.9818}}\\ 
		&FU-PSGAN & \textbf{\textcolor{blue}{1.2411}} & \textbf{\textcolor{green}{0.9869}} & 0.9865 & \textbf{\textcolor{green}{1.2907}} & \textbf{\textcolor{green}{0.9864}} & 0.0104 & \textbf{\textcolor{blue}{0.0149}} & 0.9749\\
		&ST-PSGAN & 1.2889 & \textbf{\textcolor{green}{0.9869}} & \textbf{\textcolor{blue}{0.9868}} & 1.3267 & 0.9857 &0.0138 & 0.0162& 0.9702 \\
		\bottomrule
	\end{tabularx}
\end{table*}

\subsection{Self-attention is not useful}
Attention mechanism plays an important role in human perception. It allows human brains to selectively concentrate on information meaningful to perceive tasks, while ignoring other irrelative information. Since it was introduced to deep learning~\cite{bahdanau2014neural},  attention mechanism has become one of the most valuable breakthroughs in the community and significantly boosts a variety of AI tasks ranging from NLP~\cite{vaswani2017attention} to CV~\cite{zhang2018self} domains. 

Among many attention models, self-attention (SA) has been reported to be able to generate high-quality images when incorporating GANs~\cite{zhang2018self}. Thus, in this paper, we explore to leverage the SA module to improve PSGAN. 
Following \cite{zhang2018self}, the non-local model~\cite{wang2018non} is adopted to introduce self-attention to our PSGAN. To be specific, the self-attention module is added into the 9th layer of the generator and the last layer of the discriminator. The experimental results are illustrated in Table~\ref{lable:differentconfigs}, from which we can see that on the QB and WV-2 images, PSGAN+SA performs much better than PSGAN+BN, however, still worse than the original PSGAN. Although the CC and sCC on WV-2 images, the Q$_4$ on GF-2 and WV-2 images are satisfactory to an extent, SA is not welcome in our models. 
\begin{table*}[!htb]
	\centering
	\caption{Performance comparisons on the test set of GF-2. The top three performances are highlighted with \textbf{\textcolor{red}{Red}}, \textbf{\textcolor{green}{Green}}, and \textbf{\textcolor{blue}{Blue}}.}.
	\label{table:compareonGF2}
	\begin{tabularx}{1\textwidth}{@{}lrXXXXX|XXX@{}}
		\toprule
		&         & SAM$\downarrow$ & CC$\uparrow$ & sCC$\uparrow$ & ERGAS$\downarrow$ & Q4$\uparrow$ & $D_{\lambda}\downarrow$ & $D_S\downarrow$ & QNR$\uparrow$ \\ 
		\midrule
		&SFIM~\cite{SFIM} & 1.5584 & 0.8721 & 0.9512 & 2.8705 & 0.8786  &0.0123 & 0.0446 & 0.9437\\
		&LMVM~\cite{LMVM} & 2.0111 & 0.9073 & 0.9365 & 2.3138 & 0.9037  &0.0022 & 0.0304 & 0.9675\\
		&LMM~\cite{LMVM} & 1.5527 & 0.8406 & 0.9450 & 3.0812 & 0.8387  &0.0151 & 0.0508 & 0.9349\\
		&HPF~\cite{HPF} & 1.5642 & 0.8776 & 0.9645 & 2.7818 & 0.8779  &0.0118 & 0.0425 & 0.9462\\
		&HPFC~\cite{HPF} & 1.7647 & 0.8852 & 0.9600 & 3.9200 & 0.8764  &0.0840 & 0.0899 & 0.8337\\
		&Brovey~\cite{brovey} & 1.3407 & 0.7990 & 0.9049 & 3.2624 & 0.8056  &0.0454 & 0.1693 & 0.7930\\
		&HCS~\cite{HCS} & 1.3407 & 0.8376 & 0.9392 & 3.2678 & 0.8296  &0.0200 & 0.0615 & 0.9197\\
		&IHS~\cite{IHS} & 1.8277 & 0.8109 & 0.9236 & 3.3495 & 0.8168  &0.0530 & 0.1617 & 0.7938\\
		&GS~\cite{GS} & 2.2288 & 0.7898 & 0.8947 & 3.4247 & 0.7861  &0.0819 & 0.1736 & 0.7587\\
		&BDSD~\cite{BDSD} & 1.8392 & 0.8791 & 0.9512 & 2.8705 & 0.8786  &0.0066 & 0.0523 & 0.9415\\ 
		\hline
		\hline
		&PNN~\cite{PNN} & 1.1899 & 0.9749 & 0.9821 & 1.2172 & 0.9946  & 0.0111 & 0.0494 & 0.9400  \\
		&PanNet~\cite{pannet} & {0.9370} & 0.9864 & {0.9889} & 0.8902 & 0.9971  & {0.0051} & {0.0128} & {0.9822}  \\
		&RED-cGAN~\cite{shao2019residual} &  \textbf{\textcolor{blue}{0.7442}} & \textbf{\textcolor{blue}{0.9909}} & \textbf{\textcolor{blue}{0.9931}} & 	0.7223 & 0.9981 & 0.0005 & \textbf{\textcolor{blue}{0.0088}} & 0.9908 \\	
		&PSGAN &0.7575 & \textbf{\textcolor{blue}{0.9909}} & 0.9929 &	\textbf{\textcolor{blue}{0.7233}} & \textbf{\textcolor{blue}{0.9980}} & \textbf{\textcolor{green}{0.0019}} & \textbf{\textcolor{red}{0.0060}} &\textbf{\textcolor{green}{0.9921}} \\ 
		&FU-PSGAN & \textbf{\textcolor{red}{0.7181}} & \textbf{\textcolor{red}{0.9915}} & \textbf{\textcolor{red}{0.9935}} & \textbf{\textcolor{red}{0.7013}} & \textbf{\textcolor{red}{0.9982}}& \textbf{\textcolor{blue}{0.0020}} & 0.0089 &\textbf{\textcolor{blue}{0.9892}} \\
		&ST-PSGAN &\textbf{\textcolor{green}{0.7300}}  & \textbf{\textcolor{green}{0.9913}} & \textbf{\textcolor{green}{0.9933}} & \textbf{\textcolor{green}{0.7084}} & \textbf{\textcolor{green}{0.9981}} & \textbf{\textcolor{red}{0.0008}} & \textbf{\textcolor{green}{0.0070}} & \textbf{\textcolor{red}{0.9922}} \\
		\bottomrule
	\end{tabularx}
\end{table*}

\begin{table*}[!htb]
	\centering
	\caption{Performance comparisons on the test set of WV-2. The top three performances are highlighted with \textbf{\textcolor{red}{Red}}, \textbf{\textcolor{green}{Green}}, and \textbf{\textcolor{blue}{Blue}}.}
	\label{table:compareonWV2}
	\begin{tabularx}{1\textwidth}{@{}lrXXXXX|XXX@{}}
		\toprule
		& & SAM$\downarrow$ & CC$\uparrow$ & sCC$\uparrow$ & ERGAS$\downarrow$ & Q4$\uparrow$ & $D_{\lambda}\downarrow$ & $D_S\downarrow$ & QNR$\uparrow$ \\ 
		\midrule
		&SFIM~\cite{SFIM} & 1.3411 & 0.9869 & 0.9892 & 3.5874 & 0.9873 & \textbf{\textcolor{green}{0.0016}} &\textbf{\textcolor{blue}{0.0048}} & \textbf{\textcolor{blue}{0.9936}}\\
		&LMVM~\cite{LMVM} & 1.5580 & 0.9895 & 0.9874 & 3.2472 & 0.9897 & 0.0024 &0.0053 & 0.9923\\
		&LMM~\cite{LMVM} & 1.5427  &0.9890  & 0.9879 & 3.2039 & 0.9898 & 0.0062 &0.0081 & 0.9857\\
		&HPF~\cite{HPF}  & 1.4367  &0.9890  & 0.9890 & 3.2330 & 0.9889 & \textbf{\textcolor{blue}{0.0017}} & 0.0049 & 0.9934\\
		&HPFC~\cite{HPF} & 2.6736  &0.9352  &0.9670  & 7.8543 & 0.9186 & 0.0144 &0.0342 & 0.9520\\
		&Brovey~\cite{brovey} &1.4023 & 0.9896 &0.9890 &3.1459 &0.9891 &0.0250 &0.0171 & 0.9583\\
		&HCS~\cite{HCS} & 1.4022 &0.9895 &0.9893 &3.1017 &0.9900 &0.0118 &0.0127 &0.9757\\
		&IHS~\cite{IHS} & 1.7003 &0.9859 &0.9895 &3.5379 &0.9890 &0.0354 &0.0245 &0.9409\\
		&GS~\cite{GS}   & 1.4448 &0.9889 &0.9878 &3.2192 &0.9879 &0.0100 &0.0166 &0.9735\\
		&BDSD~\cite{BDSD}&1.6422 &0.9886 &0.9924 &3.3940 &0.9905 &0.0019 &0.0052 &0.9929\\ 
		\hline
		\hline
		&PNN~\cite{PNN} & 1.4746 & 0.9955 & 0.9950 & 2.1630 & 0.9951  & 0.0086 & 0.0120 & 0.9795  \\
		&PanNet~\cite{pannet} & {0.9810} & 0.9966 & {0.9966} & 1.8530 & \textbf{\textcolor{blue}{0.9964}}  & {0.0044} & {0.0103} & {0.9854}  \\
		&RED-cGAN~\cite{shao2019residual} & \textbf{\textcolor{green}{0.8910}} & \textbf{\textcolor{green}{0.9973}} & \textbf{\textcolor{green}{0.9974}} & \textbf{\textcolor{blue}{1.6639}} & \textbf{\textcolor{green}{0.9970}} & \textbf{\textcolor{red}{0.0013}} & 0.0049 & \textbf{\textcolor{green}{0.9938}} \\
		&PSGAN\ & 0.9127 & \textbf{\textcolor{green}{0.9973}} & \textbf{\textcolor{red}{0.9975}} &	\textbf{\textcolor{green}{1.6452}} & \textbf{\textcolor{red}{0.9971}} & 0.0021 & \textbf{\textcolor{green}{0.0045}} & 0.9934 \\ 
		&FU-PSGAN \ & \textbf{\textcolor{red}{0.8855}} & \textbf{\textcolor{red}{0.9974}} & \textbf{\textcolor{green}{0.9974}} & \textbf{\textcolor{red}{1.6319}} & \textbf{\textcolor{red}{0.9971}}& \textbf{\textcolor{green}{0.0016}} & \textbf{\textcolor{red}{0.0038}} &\textbf{\textcolor{red}{0.9947}} \\	
		&ST-PSGAN & \textbf{\textcolor{blue}{0.9280}} & \textbf{\textcolor{blue}{0.9972}} & \textbf{\textcolor{blue}{0.9973}} & 1.6872 & \textbf{\textcolor{green}{0.9970}} & 0.0018 & 0.0085 & 0.9897 \\
		\bottomrule
	\end{tabularx}
\end{table*}
\subsection{Two-stream is better than stacking}
\label{subsec:twostream}
We present two variants for our PSGAN, i.e., FU-PSGAN and ST-PSGAN. PSGAN and FU-PSGAN share a similar structure that both have two-stream inputs as described in Section~\ref{subsubsec:twostreamG}. ST-PSGAN has only one input branch that accepts stacked PAN and up-sampled MS as input. Most previous works adopt a one branch design similar to ST-PSGAN, such as PNN~\cite{PNN} and PanNet~\cite{pannet}, and ignore the two-stream solution. To achieve a better performance, we evaluate different structures of PSGAN and report quantitative results in \Cref{table:compareonQB,table:compareonGF2,table:compareonWV2} with all metrics including non-reference ones are given. It should be noted that all non-reference measurements are calculated under the full-scale image setting. From these tables, we can observe that the stacking strategy, i.e. ST-PSGAN is the worst among the three PSGANs in almost all cases except for on the GF-2 images (see Table~\ref{table:compareonGF2}) where it obtains the second best results. Generally speaking, the two-stream strategy is better than stacking, and the models built with the two-stream idea are expected to achieve better performances. FU-PSGAN reaches the top performance on GF-2 images (see Table~\ref{table:compareonGF2}) and obtains satisfactory results on WV-2 (see Table~\ref{table:compareonWV2}). On QB set, it is inferior to PSGAN, however, still better than ST-PSGAN in terms of all metrics except for sCC. The three PSGANs generalize well to full-scale images. Although ST-PSGAN achieves the best $D_{\lambda}$ and QNR on GF-2 images, it still lags behind the other two PSGANs on QB and WV-2 images. FU-PSGAN performs the best on WV-2, however, the worst on GF-2 images. PSGAN is superior to the other variants on QB with the lowest $D_{\lambda}$ and $D_S$ and the highest QNR.

\begin{figure*}[!htb]\vspace{-10pt}
	\centering
	\subfigure[PAN]{\label{fig:results_of_qb_LMM}
		\includegraphics[width=4.4cm]{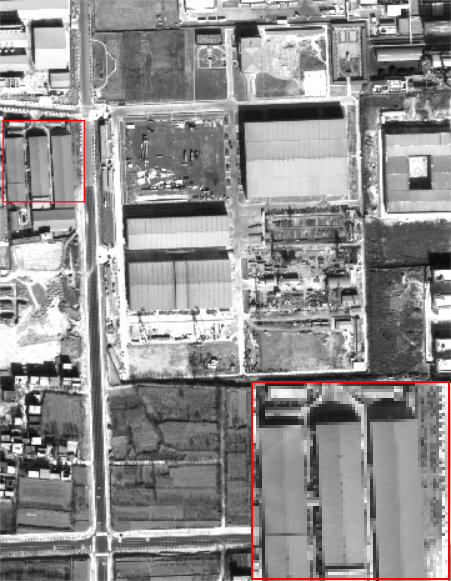}} \hspace{-8pt}\vspace{-8pt}
	\subfigure[LR MS]{\label{fig:results_of_qb_HPFC}
		\includegraphics[width=4.4cm]{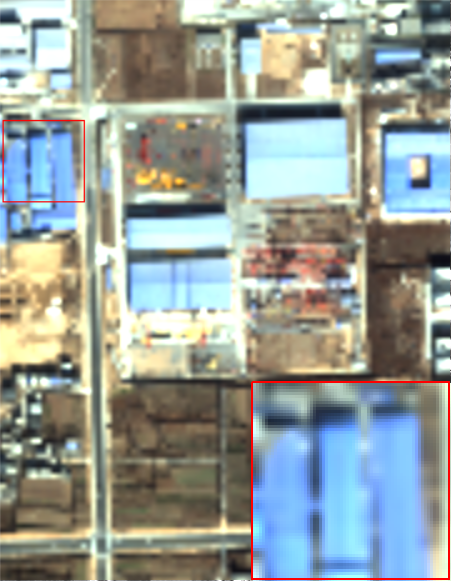}} \hspace{-8pt}
	\subfigure[SFIM\cite{SFIM}]{\label{fig:results_of_qb_SFIM}
		\includegraphics[width=4.4cm]{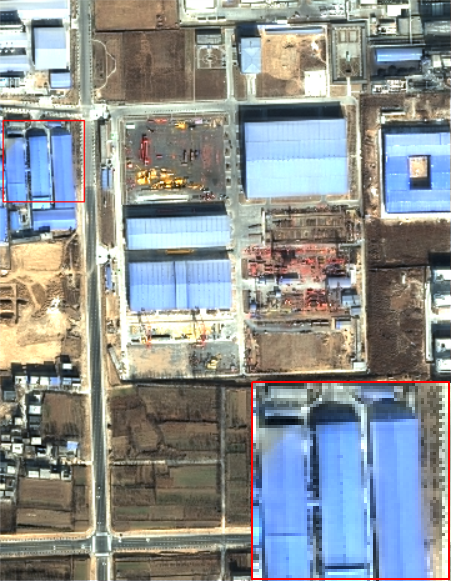}} \hspace{-8pt}
	\subfigure[LMVM\cite{LMVM}]{\label{fig:results_of_qb_LMVM}
		\includegraphics[width=4.4cm]{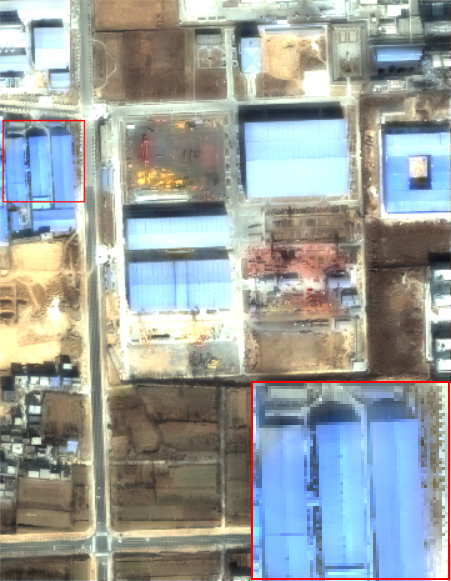}} \\
	\subfigure[HPF\cite{HPF}]{\label{fig:results_of_qb_HPF}
		\includegraphics[width=4.4cm]{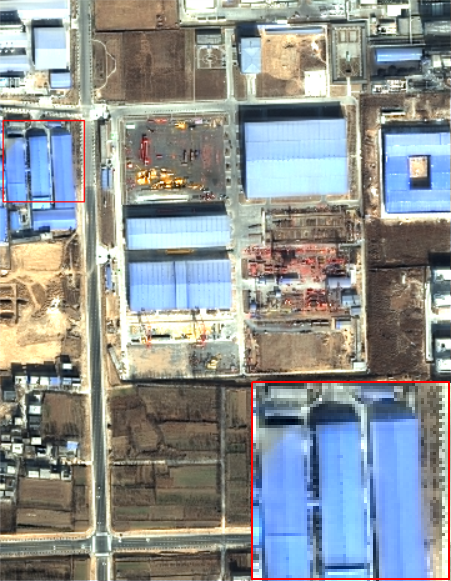}} \hspace{-8pt}\vspace{-8pt}
	\subfigure[Brovey\cite{brovey}]{\label{fig:results_of_qb_Brovey}
		\includegraphics[width=4.4cm]{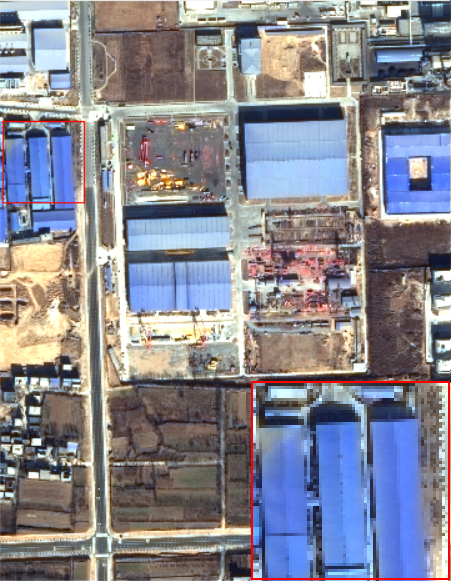}} \hspace{-8pt}
	\subfigure[HCS\cite{HCS}]{\label{fig:results_of_qb_HCS}
		\includegraphics[width=4.4cm]{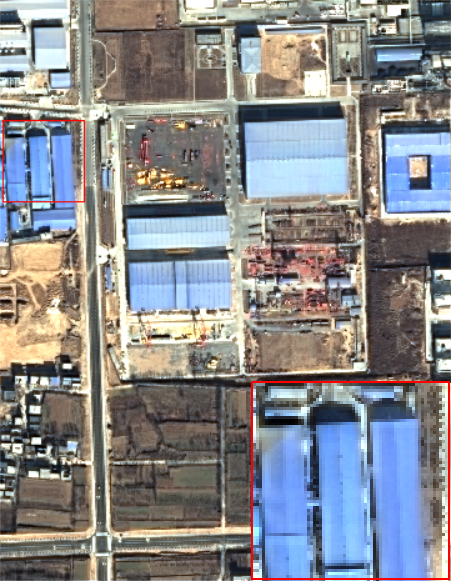}} \hspace{-8pt}
	\subfigure[IHS\cite{AIHS}]{\label{fig:results_of_qb_IHS}
		\includegraphics[width=4.4cm]{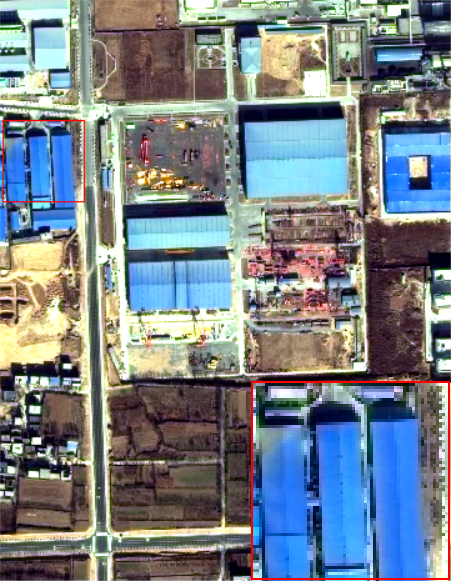}} \\
	\subfigure[GS\cite{GS}]{\label{fig:results_of_qb_GS}
		\includegraphics[width=4.4cm]{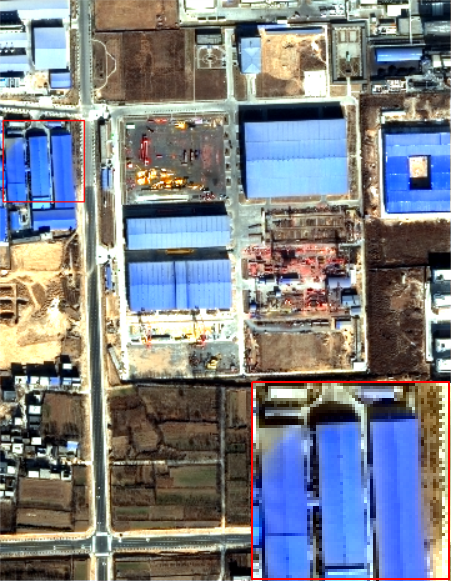}} \hspace{-8pt}\vspace{-8pt}
	\subfigure[BDSD\cite{BDSD}]{\label{fig:results_of_qb_BDSD}
		\includegraphics[width=4.4cm]{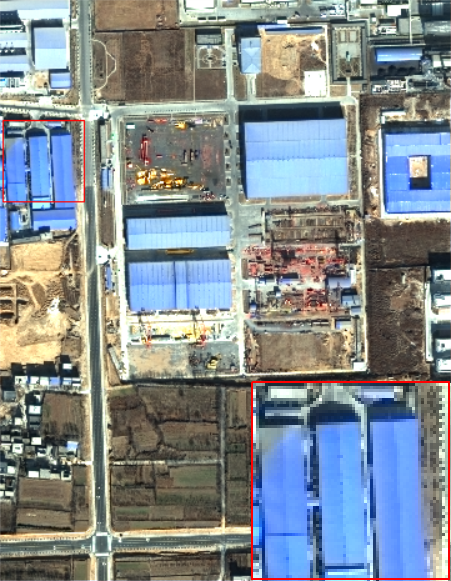}} \hspace{-8pt}
	\subfigure[PNN\cite{PNN}]{\label{fig:results_of_qb_PNN}
		\includegraphics[width=4.4cm]{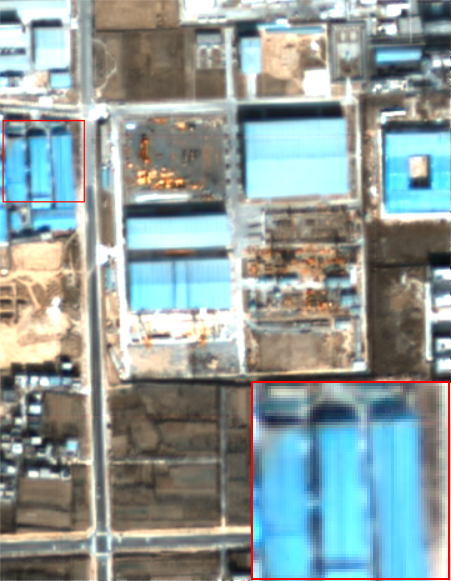}} \hspace{-8pt}
	\subfigure[PanNet\cite{pannet}]{\label{fig:results_of_qb_PanNet}
		\includegraphics[width=4.4cm]{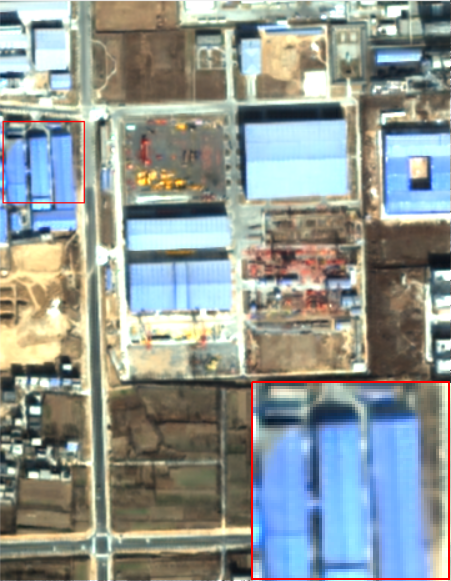}} \\
	\subfigure[PSGAN]{\label{fig:results_of_qb_PSGAN}
		\includegraphics[width=4.4cm]{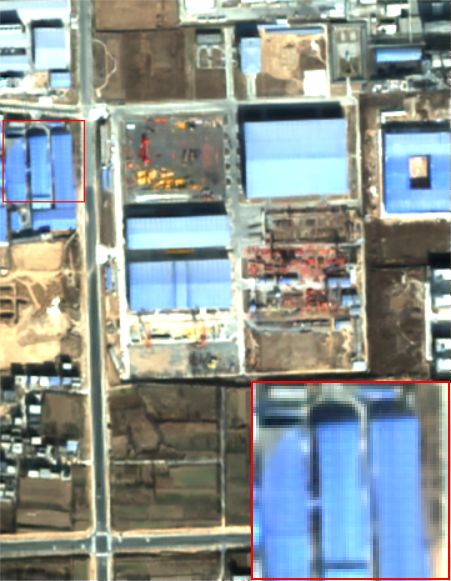}} \hspace{-8pt}\vspace{-8pt}
	\subfigure[FU-PSGAN]{\label{fig:results_of_qb_FUPSGAN}
		\includegraphics[width=4.4cm]{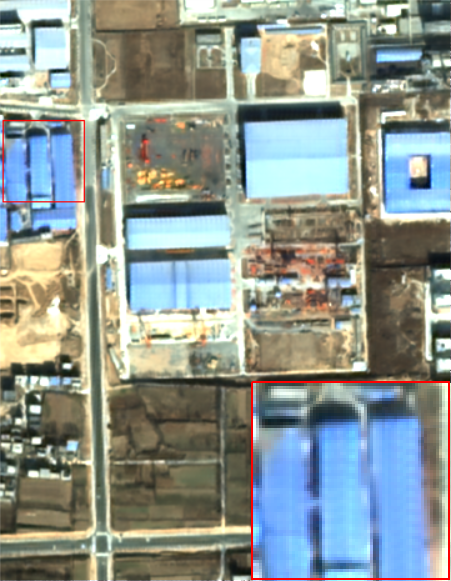}} \hspace{-8pt}
	\subfigure[ST-PSGAN]{\label{fig:results_of_qb_STPSGAN}
		\includegraphics[width=4.4cm]{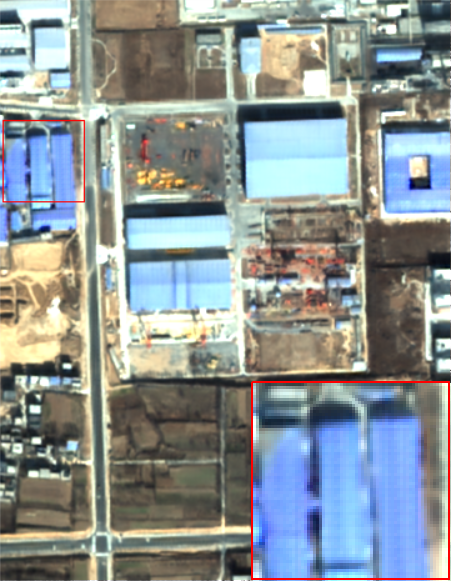}} \hspace{-8pt}
	\subfigure[GT]{\label{fig:results_of_qb_GT}
		\includegraphics[width=4.4cm]{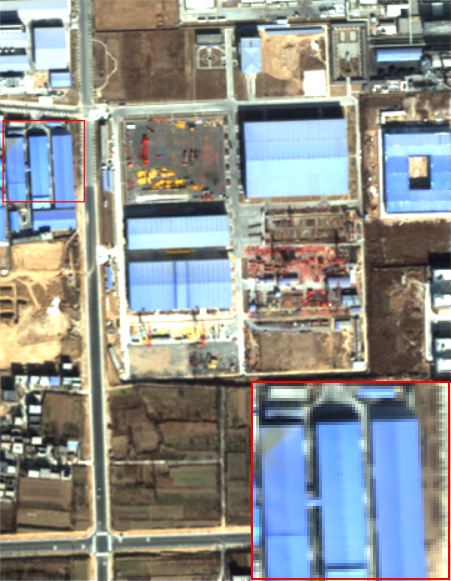}} \\
	\caption{Visual comparison on QB images. Images are displayed in RGB combination. All images have the same size of $465\times 360$ pixels.}
	\label{fig:results_of_qb}
\end{figure*}

\begin{figure*}[!htb]\vspace{-10pt}
	\centering
	\subfigure[PAN]{\label{fig:results_of_gf2_LMM}
		\includegraphics[width=4.4cm]{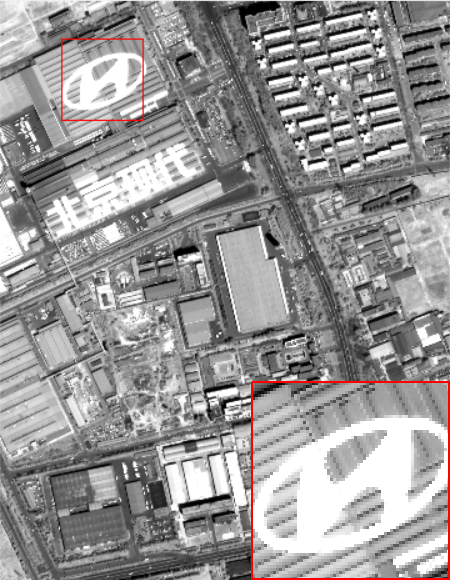}} \hspace{-8pt}\vspace{-8pt}	
	\subfigure[LR MS]{\label{fig:results_of_gf2_HPFC}
		\includegraphics[width=4.4cm]{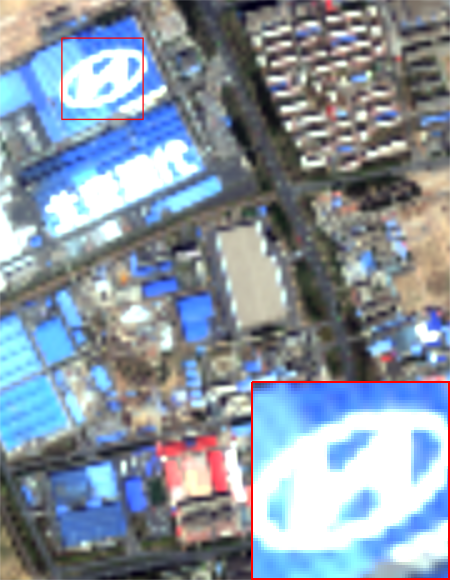}} \hspace{-8pt}
	\subfigure[SFIM\cite{SFIM}]{\label{fig:results_of_gf2_SFIM}
		\includegraphics[width=4.4cm]{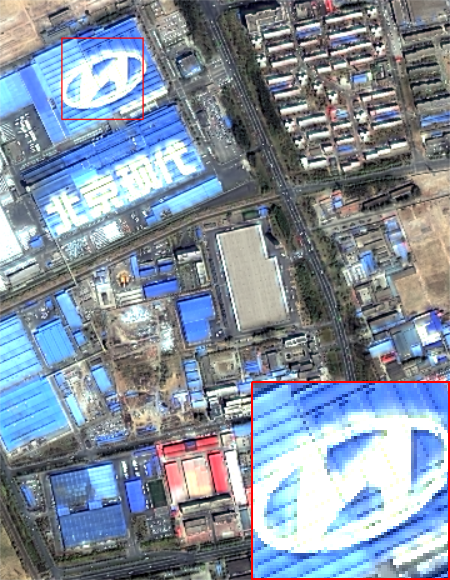}} \hspace{-8pt}
	\subfigure[LMVM\cite{LMVM}]{\label{fig:results_of_gf2_LMVM}
		\includegraphics[width=4.4cm]{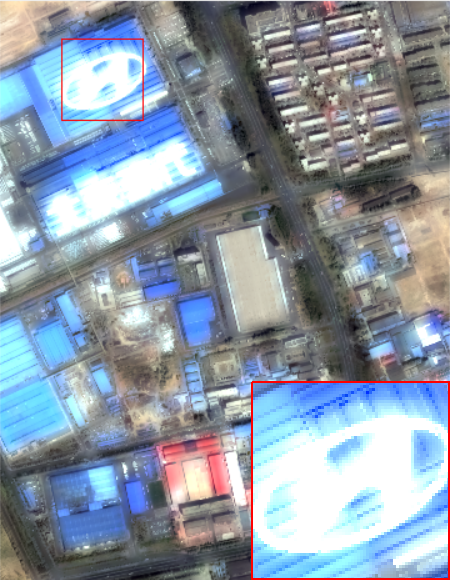}} \hspace{-8pt}\\
	\subfigure[HPF\cite{HPF}]{\label{fig:results_of_gf2_HPF}
		\includegraphics[width=4.4cm]{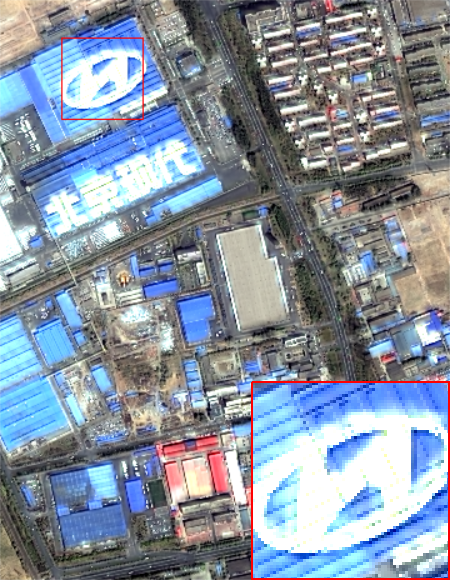}} \hspace{-8pt}\vspace{-8pt}
	\subfigure[Brovey\cite{brovey}]{\label{fig:results_of_gf2_Brovey}
		\includegraphics[width=4.4cm]{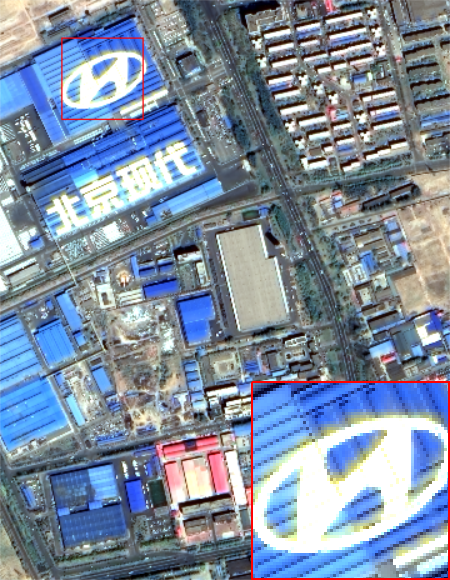}} \hspace{-8pt}
	\subfigure[HCS\cite{HCS}]{\label{fig:results_of_gf2_HCS}
		\includegraphics[width=4.4cm]{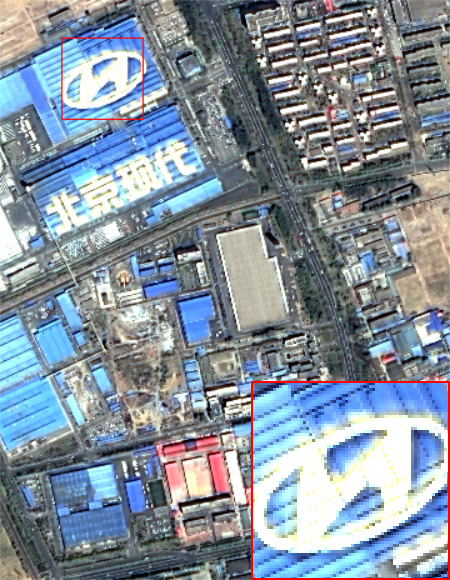}} \hspace{-8pt}
	\subfigure[IHS\cite{AIHS}]{\label{fig:results_of_gf2_IHS}
		\includegraphics[width=4.4cm]{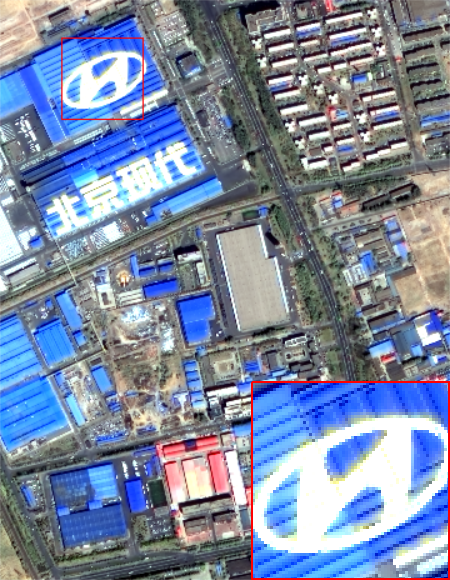}} \\
	\subfigure[GS\cite{GS}]{\label{fig:results_of_gf2_GS}
		\includegraphics[width=4.4cm]{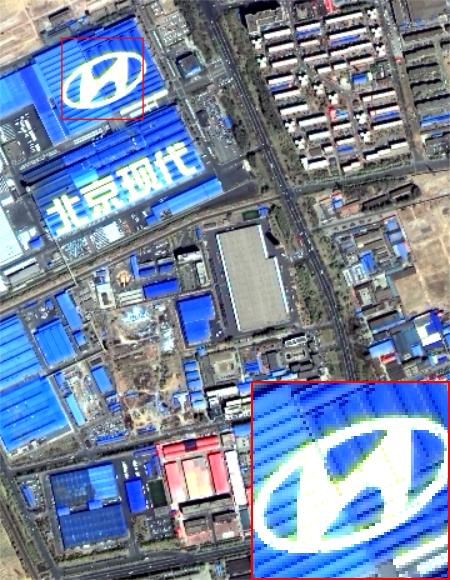}} \hspace{-8pt}\vspace{-8pt}
	\subfigure[BDSD\cite{BDSD}]{\label{fig:results_of_gf2_BDSD}
		\includegraphics[width=4.4cm]{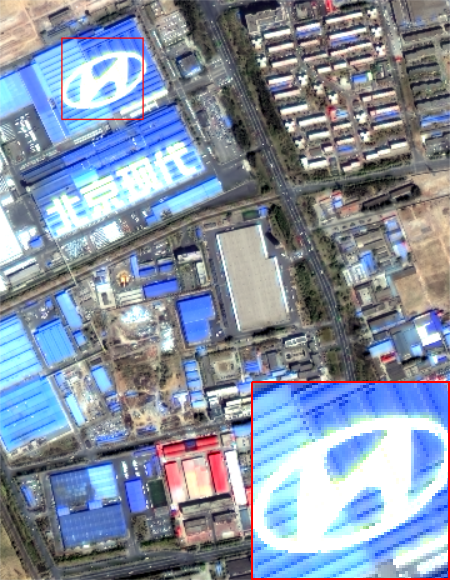}} \hspace{-8pt}
	\subfigure[PNN\cite{PNN}]{\label{fig:results_of_gf2_PNN}
		\includegraphics[width=4.4cm]{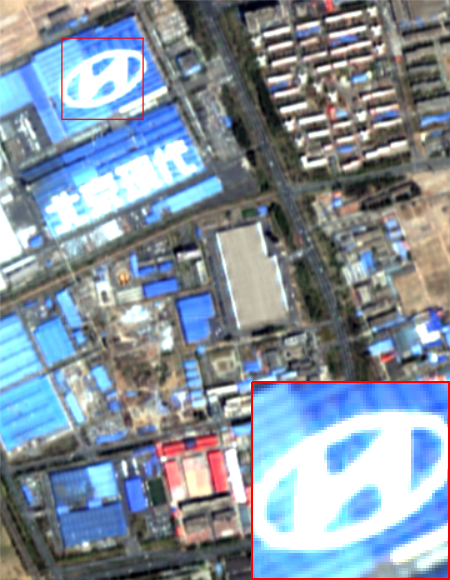}} \hspace{-8pt}
	\subfigure[PanNet\cite{pannet}]{\label{fig:results_of_gf2_PanNet}
		\includegraphics[width=4.4cm]{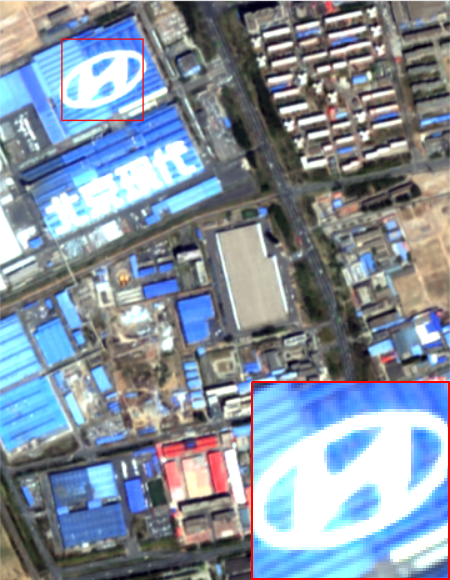}} \\
	\subfigure[PSGAN]{\label{fig:results_of_gf2_PSGAN}
		\includegraphics[width=4.4cm]{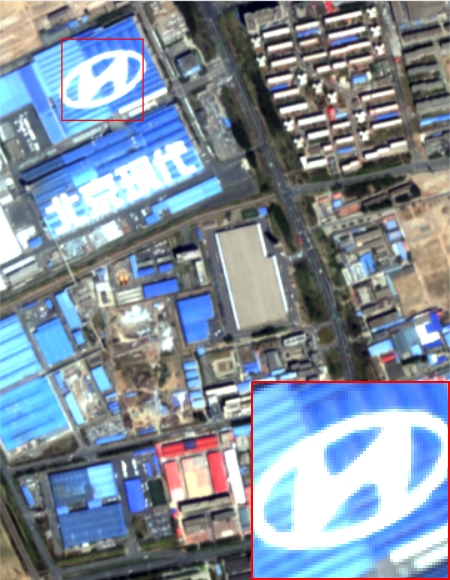}} \hspace{-8pt}\vspace{-8pt}
	\subfigure[FU-PSGAN]{\label{fig:results_of_gf2_FUPSGAN}
		\includegraphics[width=4.4cm]{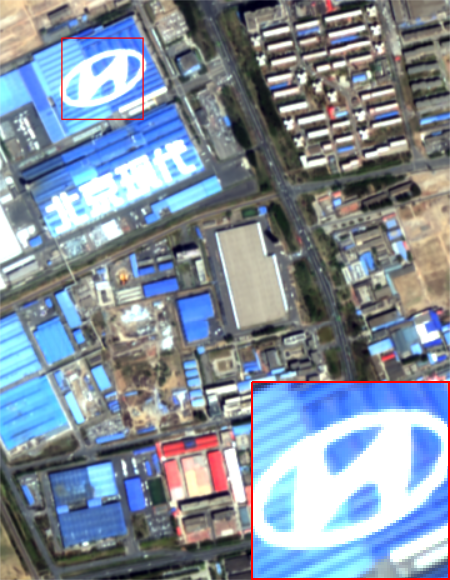}} \hspace{-8pt}
	\subfigure[ST-PSGAN]{\label{fig:results_of_gf2_STPSGAN}
		\includegraphics[width=4.4cm]{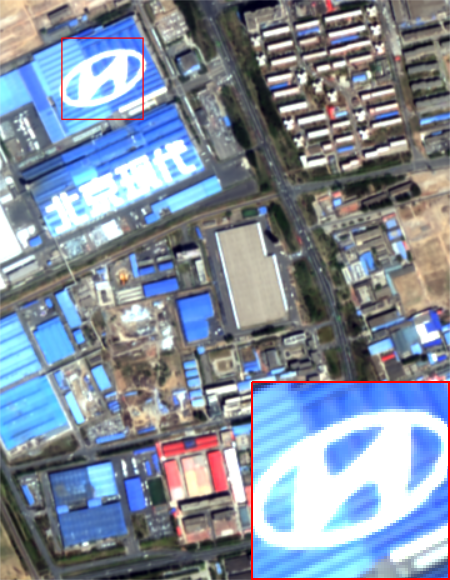}} \hspace{-8pt}
	\subfigure[GT]{\label{fig:results_of_gf2_GT}
		\includegraphics[width=4.4cm]{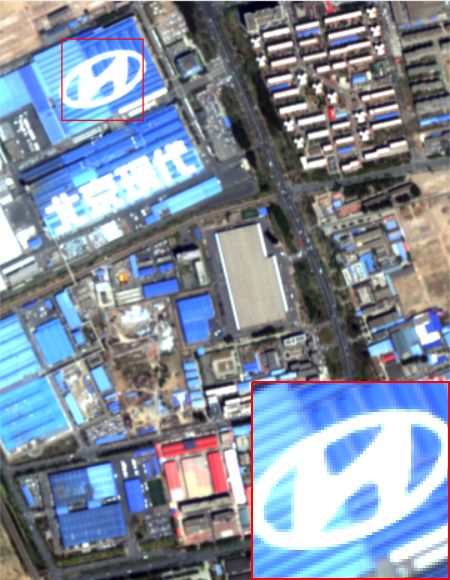}} \\
	\caption{Visual comparison on GF-2 images. Images are displayed in RGB combination. All images have the same size of $465\times 360$ pixels.}
	\label{fig:results_of_gf2}
\end{figure*}

Although the quantitative measures vary in terms of numerical metrics, the visual perceptions of them are very similar, as shown in \cref{fig:results_of_qb_PSGAN,fig:results_of_qb_FUPSGAN,fig:results_of_qb_STPSGAN}, \cref{fig:results_of_gf2_PSGAN,fig:results_of_gf2_FUPSGAN,fig:results_of_gf2_STPSGAN}, and \cref{fig:wv2-PSGAN,fig:wv2-FUPSGAN,fig:wv2-STPSGAN}. In \cref{fig:results_of_qb_PSGAN,fig:results_of_qb_FUPSGAN,fig:results_of_qb_STPSGAN} and  \cref{fig:results_of_gf2_PSGAN,fig:results_of_gf2_FUPSGAN,fig:results_of_gf2_STPSGAN}, all of them have faithful colors and spatial details to the ground truth images. \cref{fig:wv2-PSGAN,fig:wv2-FUPSGAN,fig:wv2-STPSGAN} are results on full-scale images. Careful inspection of them indicates that FU-PSGAN is the best among the three on WV-2 images, which is consistent with the quantitative results in Table~\ref{table:compareonWV2}. 

\subsection{Comparison with other pan-sharpening methods}
In this subsection, we compare the proposed PSGAN and its two improved variations, i.e. FU-PSGAN and ST-PSGAN with twelve widely used pan-sharpening techniques, including ten traditional methods: SFIM~\cite{SFIM}, LMVM~\cite{LMVM}, LMM~\cite{LMVM}, HPF~\cite{HPF}, HPFC\cite{HPF}, Brovey~\cite{brovey}, HCS~\cite{HCS}, IHS~\cite{AIHS}, GS~\cite{GS}, BDSD~\cite{BDSD}, and three deep learning based methods: PNN~\cite{PNN}, PanNet~\cite{pannet}, and RED-cGAN\cite{shao2019residual}. \Cref{table:compareonQB,table:compareonGF2,table:compareonWV2} list the quantitative evaluations on the three datasets. \Cref{table:compareonQB,table:compareonGF2} report quality indexes of all comparison methods. It can be seen deep models achieve surprisingly good performances and are superior to traditional methods in most cases. PanNet~\cite{pannet} is a successful method with very promising results. It obtains the best SAM on QB images and generalizes well to full-scale images, which is supported by its optimal non-reference metrics. As a pioneering deep model, PNN~\cite{PNN} proves the effectiveness of applying deep neural networks to pan-sharpening tasks. Although PNN has the lowest spectral quality on the QB dataset, it works well on the GF-2 images with remarkable SAM surpassing all traditional methods. The proposed PSGAN obtains the best metrics on the QB images except for the SAM indicator. On GF-2 dataset, PSGAN and its variants show superiority performance than all other methods. Especially, increasing the spatial resolution of MS images using CNN networks, i.e., \mbox{FU-PSGAN} gains the best performance. Stacking the MS and PAN images together to perform pan-sharpening achieve the second-best place on the GF-2 dataset, however it falls behind the other two PSGAN models. \Cref{table:compareonWV2} presents the quantitative results of deep models on WV-2 images. As can be observed, our models still perform better than PNN~\cite{PNN} and PanNet~\cite{pannet}. Especially, FU-PSGAN achieves the best results on this dataset with the highest SAM, CC, ERGAS, and Q$_4$, and slightly worse sCC than PSGAN.

\begin{table*}[!htb]
	\centering
	\caption{Computational costs and number of parameters of different models on the test sets. Note that the pan-sharpened images are with sizes of around $3000\times 2048\times 4$, we give average time on them.}
	\label{lable:computationalcosts}
	\begin{tabularx}{1\textwidth}{@{}lrXXX@{}}
		\toprule
		Processor & Method & GFLOPS & Time (s) & $\sharp$Params  \\ 
		\midrule
		\multirow{10}{*}{\makecell{Intel Core i7-7700HQ CPU@2.80GHz}} 
		&SFIM~\cite{SFIM}     & -         & 3.54  & - \\ 
		&LMVM~\cite{LMVM}     & -         & 22.60 & - \\  
		&LMM~\cite{LMVM}      & -         & 3.57  & - \\ 
		&HPF~\cite{HPF}      & -         & 3.66  & - \\
		&HPFC~\cite{HPF}     & -         & 3.10  & - \\
		&Brovey~\cite{brovey}   & -         & 0.41  & - \\
		&HCS~\cite{HCS}      & -         & 4.13  & - \\
		&IHS~\cite{IHS}      & -         & 0.39  & - \\
		&GS~\cite{GS}       & -         & 3.90  & - \\
		&BDSD~\cite{BDSD}     & -         & 40.36 & - \\
		\midrule
		\multirow{5}{*}{\makecell{NVIDIA GeForce RTX 2080Ti}} 
		&PNN~\cite{PNN}       & $\sim$ 84   & 0.43  & $\sim$ 0.080 M \\ 
		&PanNet~\cite{pannet}   & $\sim$ 80   & 0.53  & $\sim$ 0.077 M \\ 
		&RED-cGAN~\cite{shao2019residual} & $\sim$ 603  & 1.35  & $\sim$ 1.90 M \\		 
		&PSGAN    & $\sim$ 402  & 1.13  & $\sim$ 1.88 M \\ 
		&PSGAN-f16  & $\sim$ 97 & 0.62  & $\sim$ 0.47 M \\
		&PSGAN-k5$\times 5$ & $\sim $ 1011 & 1.52 & $\sim $ 4.74 M \\
		&FU-PSGAN & $\sim$ 392  & 1.07  & $\sim$ 1.89 M\\
		&ST-PSGAN & $\sim$ 351  & 0.98  & $\sim$ 1.77 M \\
		\bottomrule
	\end{tabularx}
\end{table*}
\subsection{Visual comparisons}

\cref{fig:results_of_qb,fig:results_of_gf2} show sample results cropped from the test site of Quickbird and GF-2 datasets, respectively. All images are displayed in true color. In~\cref{fig:results_of_qb,fig:results_of_gf2}, LMVM~\cite{LMVM} and PNN~\cite{PNN} tend to blur images with very poor visual quality (see \cref{fig:results_of_qb_LMVM,fig:results_of_gf2_LMVM,fig:results_of_qb_PNN,fig:results_of_gf2_PNN}). SFIM~\cite{SFIM}, Brovey~\cite{brovey}, HCS~\cite{HCS}, IHS~\cite{AIHS}, GS~\cite{GS}, and BDSD~\cite{BDSD} perform spatial information injection efficiently and produce results with clean high-frequency details almost identical to the PAN images. However, they suffer from severe spectral distortions, especially Brovey (\cref{fig:results_of_qb_Brovey}), IHS (\cref{fig:results_of_qb_IHS}), and GS (\cref{fig:results_of_qb_GS}) methods, the colors of which are darker than GT and MS images on the QB test set. Brovey (\cref{fig:results_of_gf2_Brovey}), HCS (\cref{fig:results_of_gf2_HCS}), IHS (\cref{fig:results_of_gf2_IHS}), GS (\cref{fig:results_of_gf2_GS}), and BDSD (\cref{fig:results_of_gf2_BDSD}) shows noticeable color distortions on the GF-2 dataset. The learning-based methods, i.e., PNN, PanNet and ours are optimized to generate images as close as to the GT images, thus they have better results when comparing with the GT images. The proposed PSGANs produce results most similar to the GTs (see \cref{fig:results_of_qb_PSGAN,fig:results_of_qb_FUPSGAN,fig:results_of_qb_STPSGAN,fig:results_of_gf2_PSGAN,fig:results_of_gf2_FUPSGAN,fig:results_of_gf2_STPSGAN}). One notable drawback of our methods as well as the other deep models is that they tend to produce smoother results than traditional ones, as can be seen from \cref{fig:results_of_qb,fig:results_of_gf2}. This is mainly because of the pixel-wise average problem~\cite{ledig2017photo} introduced by the loss function involving averaging operation such as $L_2$. This phenomenon is frequently observed in image enhancement tasks. One possible solution to this is using perceptual losses~\cite{johnson2016perceptual} which will be considered in our future work.

\subsection{Experiments on full resolution images}
\begin{figure*}
	\centering
	\subfigure[PAN]{\label{fig:wv2-PAN}
		\begin{minipage}[b]{0.23\linewidth}
			\includegraphics[width=4.5cm]{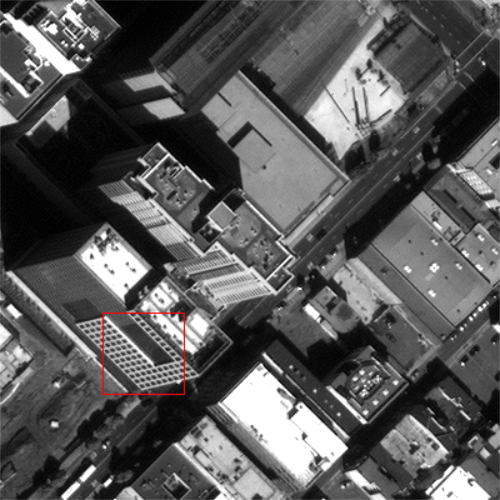}\vspace{2pt}
			\includegraphics[width=4.5cm]{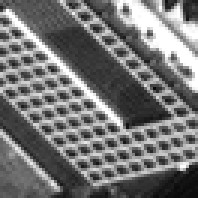}
		\end{minipage}}
		\hfill
	\subfigure[LR MS]{\label{fig:wv2-LRMS}
		\begin{minipage}[b]{0.23\linewidth}
			\includegraphics[width=4.5cm]{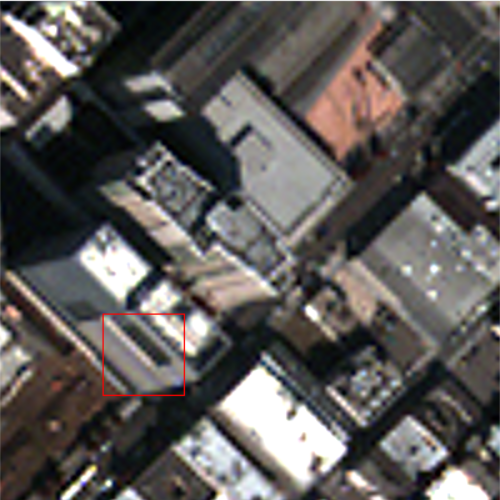}\vspace{2pt}
			\includegraphics[width=4.5cm]{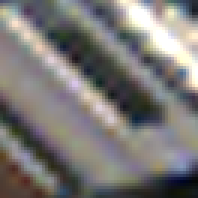}
		\end{minipage}}
		\hfill
	\subfigure[SFIM~\cite{SFIM}]{\label{fig:wv2-SFIM}
		\begin{minipage}[b]{0.23\linewidth}
			\includegraphics[width=4.5cm]{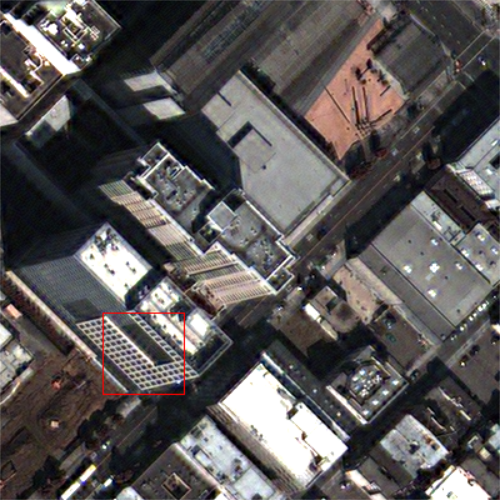}\vspace{2pt}
			\includegraphics[width=4.5cm]{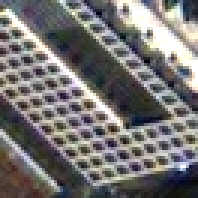}
		\end{minipage}}
		\hfill
	\subfigure[PNN~\cite{PNN}]{\label{fig:wv2-PNN}
		\begin{minipage}[b]{0.23\linewidth}
			\includegraphics[width=4.5cm]{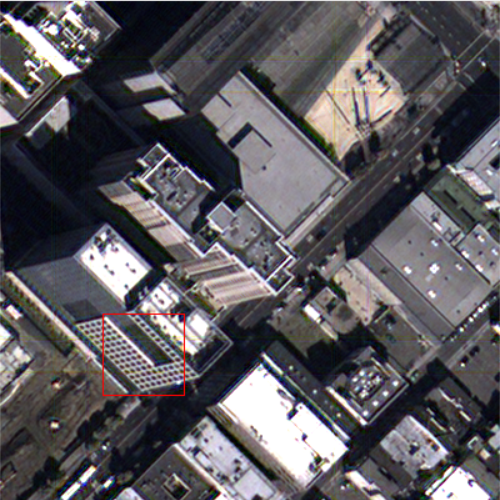}\vspace{2pt}
			\includegraphics[width=4.5cm]{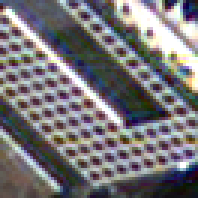}
		\end{minipage}}
	\subfigure[PanNet~\cite{pannet}]{\label{fig:wv2-PANNET}
		\begin{minipage}[b]{0.23\linewidth}
			\includegraphics[width=4.5cm]{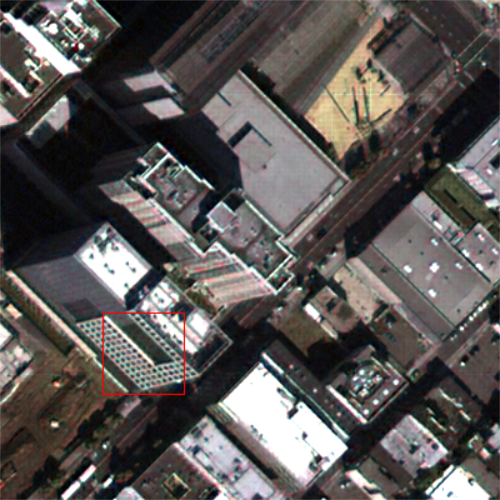}\vspace{2pt}
			\includegraphics[width=4.5cm]{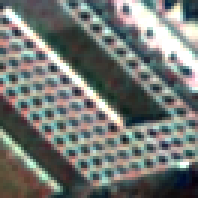}
		\end{minipage}}
		\hfill
	\subfigure[PSGAN]{\label{fig:wv2-PSGAN}
		\begin{minipage}[b]{0.23\linewidth}
			\includegraphics[width=4.5cm]{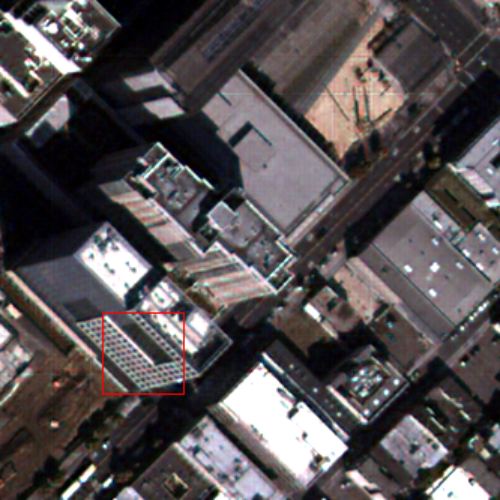}\vspace{2pt}
			\includegraphics[width=4.5cm]{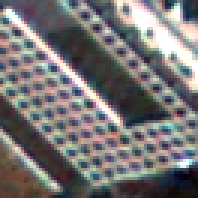}
		\end{minipage}}
		\hfill
	\subfigure[FU-PSGAN]{\label{fig:wv2-FUPSGAN}
		\begin{minipage}[b]{0.23\linewidth}
			\includegraphics[width=4.5cm]{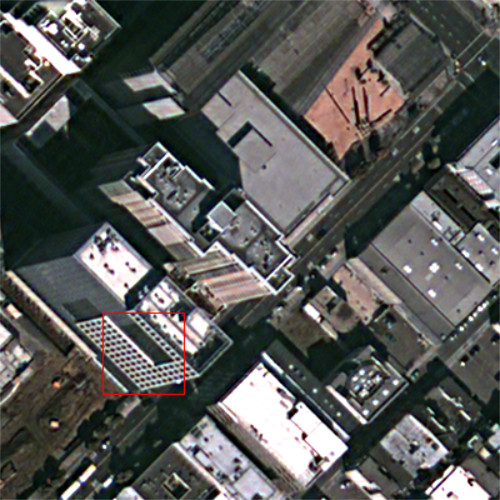}\vspace{2pt}
			\includegraphics[width=4.5cm]{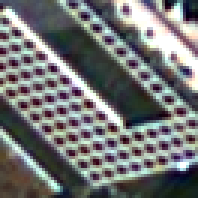}
		\end{minipage}}
		\hfill
	\subfigure[ST-PSGAN]{\label{fig:wv2-STPSGAN}
		\begin{minipage}[b]{0.23\linewidth}
			\includegraphics[width=4.5cm]{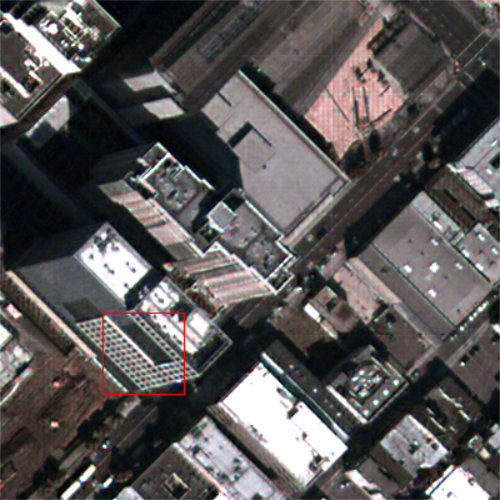}\vspace{2pt}
			\includegraphics[width=4.5cm]{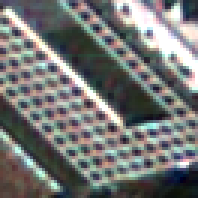}
		\end{minipage}}
	\caption{Example results on WV-2 images ($400\times 400$ pixels). Displayed in RGB channels.}
	\label{fig:full_scale_results_of_WV2}
\end{figure*}

We also evaluate our models on full-scale images without down-sampling them. It should be noted that, under this setting, there will be no target images available for training. Considering that generalization ability across scales is the main concern in this experiment, we directly apply the optimized networks to the original PAN and MS images to produce the desired HR MS images. For quantitative evaluation, we calculate non-reference indexes as described in \cref{equ:dambda,equ:ds,equ:qnr} for each pan-sharpened image and report results on the right sides of \cref{table:compareonQB,table:compareonGF2,table:compareonWV2}. As can be seen, the proposed PSGANs generalize well to the full-scale images. They obtain competitive performance on the three datasets. Especially, FU-PSGAN achieves the best results on WV-2 images, some typical samples are represented in \cref{fig:full_scale_results_of_WV2} which clearly shows appealing results of FU-PSGAN.
\subsection{Computational time}
We test the computational time of ours and the other comparison methods. All traditional methods are implemented using Matlab and run on an Intel Core i7-7700HQ CPU, and deep models are implemented in PyTorch and tested on a single NVIDIA GeForce RTX 2080Ti GPU. The computational times are computed on the test sets of the three datasets. We give average time on the test sets for each method. Traditional methods are much faster than deep learning models. IHS~\cite{IHS} and Brovey~\cite{brovey} are the fastest ones. It takes less about 0.5 seconds for them to produce pan-sharpened images with sizes of about $3000\times2048\times4$. SFIM~\cite{SFIM}, LMM~\cite{LMVM}, HPF~\cite{HPF}, HPFC~\cite{HPF}, and GS~\cite{GS} have almost the same time. They spend about $3\sim 4$ seconds to pan-sharpen one image. HCS~\cite{HCS} takes a little longer, it takes more than 4 seconds to process one image. LMVM~\cite{LMVM} and BDSD~\cite{BDSD} are among the most time-consuming pan-sharpening. They spend 22 and 40 seconds, respectively, for generating one image. Beneficial from the advance of GPU architectures, deep learning based models are satisfactory. PNN~\cite{PNN} and PanNet~\cite{pannet} take about 0.43 and 0.53 seconds to process one image. It costs about 1.13, 1.07, and 0.98 seconds for PSGAN, FU-PSGAN, and ST-PSGAN to pan-sharpen one image. Our models are slower than PNN and PanNet because we have deeper architectures than them. FU-PSGAN performs a bit faster than PSGAN because it has de-convolution operations so that the input size is smaller. RED-cGAN~\cite{shao2019residual} takes longer than PSGAN, because it has more parameters. 

\section{Conclusion}
\label{sec:conclusion}
In this paper, we have proposed PSGANs for solving the task of image pan-sharpening and conducted extensive experiments on Quckbird, GaoFen-2, and Worldview-2 images. The experiments demonstrate that the PSGANs are effective in generating high-quality pan-sharpened images with fine spatial details and high-fidelity spectral information under both low-scale and full-scale image settings, and are superior to many popular pan-sharpening approaches. Furthermore, we evaluate several designs including two-stream input, stacking input, batch normalization layer, and attention mechanism to find the optimal solution for the pan-sharpening task. We find that the two-stream architecture is normally better than the stacking strategy, and the batch normalization layer and the self-attention module are not welcome in pan-sharpening. We suggest removing them from networks when designing pan-sharpening models. 

In our future work, we will focus on unsupervised learning for pan-sharpening. Although remarkable results have been achieved by PSGANs, their generalization ability to full-scale images is still underdeveloped. We intend to solve this problem under unsupervised learning framework and optimize the models using only original PAN and MS images without any preprocessing steps. 


\ifCLASSOPTIONcaptionsoff
  \newpage
\fi



%
\bibliographystyle{IEEEbib}
\bibliography{transref}

\end{document}